%% file: main.tex
\pdfoutput=1

\documentclass[11pt]{article}

\usepackage[final]{acl}

\usepackage{times}
\usepackage{latexsym}

\usepackage[T1]{fontenc}

\usepackage[utf8]{inputenc}

\usepackage{microtype}

\usepackage{inconsolata}

\usepackage{graphicx}

\usepackage{booktabs}
\usepackage{amsmath}
\usepackage{multirow}
\usepackage{cleveref}
\usepackage{enumitem}
\usepackage{listings}
\usepackage{pgfplots}
\usepackage{tcolorbox}
\pgfplotsset{compat=1.13}
\definecolor{decentgrey}{RGB}{232,232,232}
\newtcbox{\pattern}{on line,colback=decentgrey,colframe=white,size=fbox,arc=3pt, box align=base, before upper=\strut, top=-2pt, bottom=-2pt, boxrule=0pt}

\usepackage{algorithm}
\usepackage{algpseudocode}
\usepackage{subcaption}

\title{CUB: Benchmarking Context Utilisation Techniques for Language Models}

\def\authorsep{\hspace{0.3em}}

\author{Lovisa Hagström\textsuperscript{1,2*} \authorsep Youna Kim\textsuperscript{3*} \authorsep Haeun Yu\textsuperscript{4} \\ \textbf{Sang-goo Lee\textsuperscript{3}} \authorsep \textbf{Richard Johansson\textsuperscript{1,2}}
\textbf{Hyunsoo Cho\textsuperscript{5$\dagger$}} \authorsep \textbf{Isabelle Augenstein\textsuperscript{4}} \medskip\\
\null\textsuperscript{1}Chalmers University of Technology \quad \null\textsuperscript{2}University of Gothenburg\\
\null\textsuperscript{3}Seoul National University \quad \null\textsuperscript{4}University of Copenhagen \quad \null\textsuperscript{5}Ewha Womans University\\
\texttt{anna9812@europa.snu.ac.kr}}

\begin{document}
\maketitle

\def\thefootnote{*}\footnotetext{Equal contribution.}\def\thefootnote{\arabic{footnote}}
\renewcommand{\thefootnote}{\fnsymbol{footnote}}
\footnotetext[2]{Dept. of AI, Institute for Multiscale Matter and Systems}

\renewcommand{\thefootnote}{\arabic{footnote}}
\setcounter{footnote}{0}

\begin{abstract}
Incorporating external knowledge is crucial for knowledge-intensive tasks, such as question answering and fact checking.
However, language models (LMs) may ignore relevant information that contradicts outdated parametric memory or be distracted by irrelevant contexts.
While many context utilisation manipulation techniques (CMTs) have recently been proposed to alleviate these issues, few have seen systematic comparison.
In this paper, we develop \textsc{CUB} (Context Utilisation Benchmark) -- the first comprehensive benchmark designed to help diagnose CMTs under diverse noisy context conditions within retrieval-augmented generation (RAG).
With this benchmark, we conduct the most extensive evaluation to date of seven state-of-the-art methods, representative of the main categories of CMTs, across three diverse datasets and tasks, applied to 11 LMs. 
Our findings expose critical gaps in current CMT evaluation practices, demonstrating the need for holistic testing.
We reveal that most existing CMTs struggle to handle the full spectrum of context types encountered in real-world RAG scenarios.
We also find that many CMTs display inflated performance on simple synthesised datasets, compared to more realistic datasets with naturally occurring samples.

\end{abstract}

\section{Introduction}\label{sec:intro}

\begin{figure}[h]
    \centering
    \includegraphics[width=1\linewidth,trim={16cm 4.2cm 17cm 0.3cm},clip]{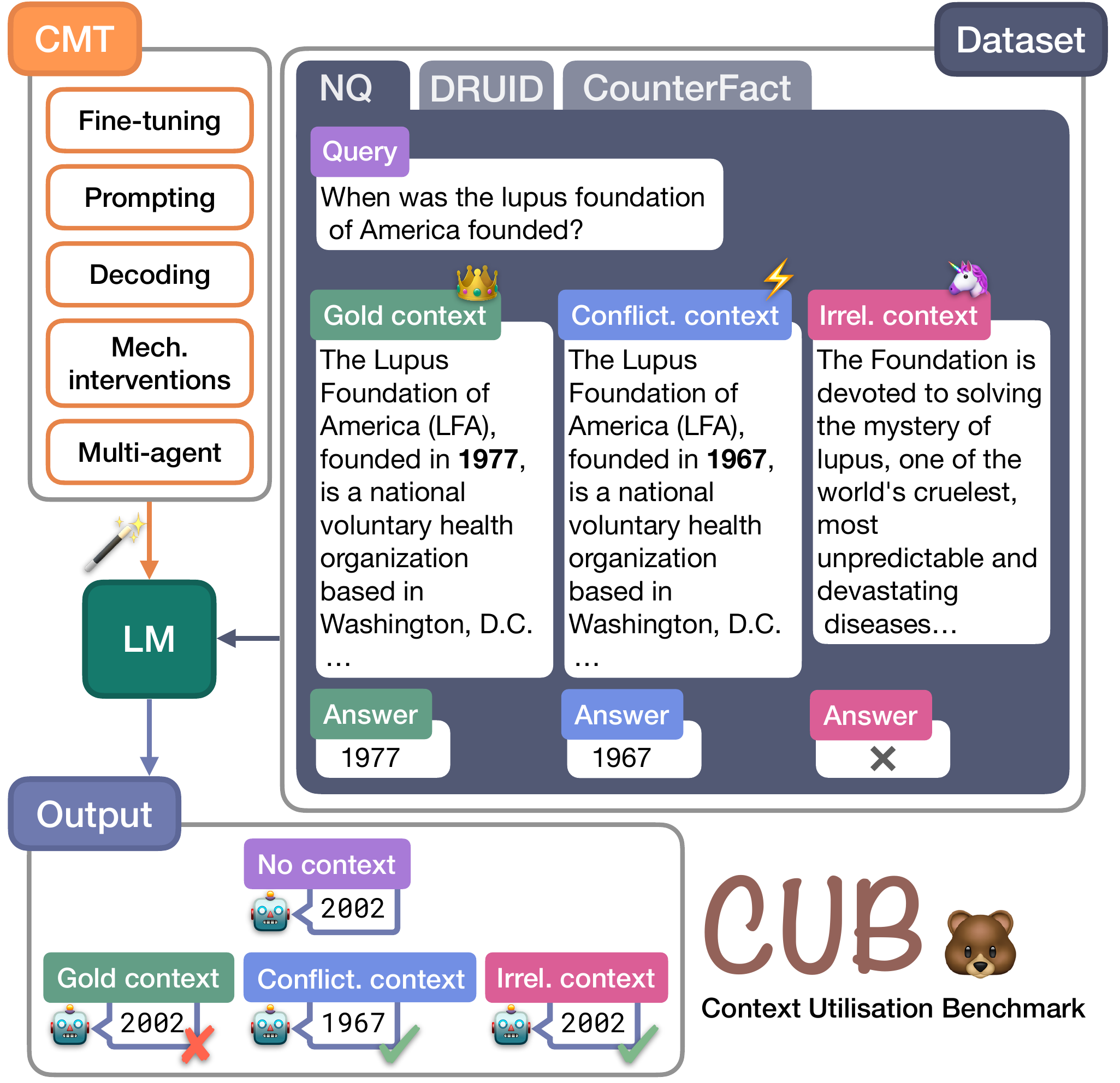}
    \caption{The Context Utilisation Benchmark. We evaluate a range of LMs under different CMTs on samples from NQ, DRUID and CounterFact for gold, conflicting and irrelevant contexts.}
    \label{fig:overview}
\end{figure}
Context utilisation is a key component of language models (LMs) used for retrieval-augmented generation (RAG), as the benefits of retrieving external information are only realised if the generative model makes adequate use of the retrieved information. While recent research has identified many benefits of augmenting LMs with retrieved information \citep{shuster-etal-2021-retrieval-augmentation,hagstrom-etal-2023-effect}, it has also identified weaknesses of LMs used for RAG, of which many are associated with context utilisation. For example, LMs can easily be distracted by irrelevant contexts \citep{llm-irrelevant-context} or ignore relevant contexts due to memory-context conflicts \citep{xu-etal-2024-knowledge-conflicts}. The robustness of LMs to irrelevant contexts is important as information retrieval systems used for RAG are not guaranteed to always retrieve relevant information. Moreover, as information may be updated to conflict with the training data of the LM, the model should prioritise the most recently updated information.

As a consequence, many different methods for increasing or suppressing LM context utilisation, henceforth referred to as \emph{CMTs} (Context utilisation Manipulation Techniques), have been proposed. The methods encompass a broad range of approaches, from different decoding methods \citep{shi-etal-2024-trusting,kim-etal-2024-adaptive} to fine-tuning methods \citep{li-etal-2023-large}, prompting \citep{prompt-survey}, multi-agent \citep{feng-etal-2024-dont, 10.5555/3692070.3692537}, and mechanistic interventions \citep{ortu-etal-2024-competition,jin-etal-2024-cutting,sun2025evaluationframeworkhighlightexplanations}. 
Many of these CMTs are developed with distinct objectives in mind, often targeting a specific aspect of context utilisation -- such as improving robustness to irrelevant contexts or enhancing faithfulness to conflicting information. 
While each method yields promising results in isolation, their evaluation is often limited to settings that closely match the method's target objective, making it unclear whether these approaches will generalise or remain effective in more realistic RAG scenarios.
To address this pressing evaluation gap, we develop the first comprehensive CMT benchmark to systematically test and compare different CMTs across datasets that represent the diversity of real-world domains and tasks (\Cref{fig:overview}).
Our \textbf{contributions} are as follows: 
\begin{itemize}[noitemsep,topsep=0pt,leftmargin=*]
    \item We develop \textsc{CUB} (Context Utilisation Benchmark) -- the first comprehensive benchmark designed to enable rigorous evaluation and comparison of CMTs under various noisy context conditions (\S\ref{sec:cub}).\footnote{Code and datasets available at \url{https://github.com/copenlu/cmt-benchmark}} \textsc{CUB} evaluates CMTs across gold, conflicting, and irrelevant contexts that capture key RAG challenges.
    \item We conduct an extensive evaluation of CMTs, assessing seven state-of-the-art methods representative of the main categories of CMTs (\S\ref{sec:cmts}) across our benchmark (\S\ref{sec:results}). The study encompasses approximately 800 experimental data points, spanning 11 LMs, 7 CMTs, 3 datasets, and 3 different context types. We believe this is the most extensive evaluation of CMTs to date.
    \item We provide insights into what CMT works best for different scenarios and identify critical areas for improvement. 
    Our analysis reveals that existing CMTs face fundamental trade-offs -- they struggle to optimise performance across all context types.
    This points to a need for next-generation CMTs that can robustly handle diverse context conditions.
\end{itemize}

\section{Related Work}

\paragraph{Context-intensive datasets}
We consider two main categories of context-intensive datasets: 1) datasets representing \emph{knowledge-intensive tasks}, i.e. tasks for which access to external context is crucial, and 2) datasets designed to \emph{diagnose} model adaptability to external knowledge. Examples of datasets representative of knowledge-intensive tasks are Natural Questions (NQ), DRUID, the KILT datasets and PubMedQA \citep{kwiatkowski-etal-2019-natural,hagstrom-etal-2025-reality,petroni-etal-2021-kilt,jin-etal-2019-pubmedqa}. 
Examples of diagnostic datasets representative of the latter category are CounterFact and ConflictQA \citep{meng2022locating,xie2024knowledgeconflict}, which contain synthesised queries based on fact triplets.
In these datasets, contexts are synthesised to induce \emph{knowledge conflicts} by promoting answers that conflict with the parametric memory of the studied LM.
These diagnostic datasets have found widespread use for work on mechanistic interpretability and the evaluation of context utilisation \citep{meng2022locating,geva-etal-2023-dissecting,ortu-etal-2024-competition,sun2025evaluationframeworkhighlightexplanations,islam2026multistepknowledgeinteractionanalysis}.

Previous work has typically evaluated different CMTs on either of the dataset categories in isolation, creating a fragmented understanding of CMT effectiveness. \textsc{CUB} bridges this critical gap by incorporating datasets representative of both knowledge-intensive tasks and diagnostic datasets, enabling the first comprehensive evaluation of CMTs across truly diverse settings.

\paragraph{CMTs}
Many context utilisation manipulation techniques have recently been proposed. Existing CMTs can be categorised into one of four main groups based on \emph{intervention level}, i.e. what aspect of the model they manipulate. 1) \emph{fine-tuning} CMTs update model parameters to modify context utilisation. For example, fine-tuning on distracting contexts was found to yield improved robustness to distracting contexts \citep{li-etal-2023-large,shen-etal-2024-assessing,yoran2024makingretrievalaugmentedlanguagemodels}. \citet{fang-etal-2024-enhancing} specifically focus on different types of retrieval noise likely to be encountered in real-world environments and develop a fine-tuning approach to handle these. 2)   \emph{prompting techniques} modify the input to the LM to improve context utilisation, representing minimally modified settings. 3) \emph{mechanistic interventions} on the LM modify certain model components at inference time to alter context utilisation. Examples involve attention modification \citep{ortu-etal-2024-competition,jin-etal-2024-cutting} and SpARe interventions \citep{zhao-etal-2025-steering}. Lastly, 4) \emph{decoding methods} involve a modified decoding approach, applied to the output logits, to manipulate context utilisation. Examples include context-aware contrastive decoding \citep{yuan-etal-2024-discerning, kim-etal-2024-adaptive, shi-etal-2024-trusting, wang2024adacadadaptivelydecodingbalance, zhao-etal-2024-enhancing} and 
lookback lens decoding \citep{chuang-etal-2024-lookback}.

Thus far, the majority of recent CMTs have been developed with standard-length contexts in mind, since the long-context setting is associated with separate challenges and approaches. For example, \citet{zhang2024middlelanguagemodelsuse} propose a method to address the \emph{lost-in-the-middle} challenge in long-context settings.

Despite significant progress in developing new CMTs, systematic evaluations remain scarce. Prior studies have largely assessed individual CMTs in isolation, likely due to the lack of a unified benchmark. In this paper, we address this gap by conducting the first comprehensive evaluation of seven representative CMTs designed for standard-length contexts across the four main categories of CMTs.

\paragraph{Benchmarks}
To our knowledge, there is not yet a benchmark specifically designed for CMTs, representing a significant research gap. The closest examples of existing benchmarks are RAG-Bench by \citet{fang-etal-2024-enhancing}, KILT by \citet{petroni-etal-2021-kilt} and AxBench by \citet{wu2025axbenchsteeringllmssimple}. However, these serve different purposes: the first evaluates the retrieval-noise robustness of LMs, the second assesses performance of RAG systems as a whole, and the latter focuses on steering techniques for LMs, emphasising safety and reliability rather than context utilisation. 
\textsc{CUB} addresses this critical infrastructure gap by creating the first comprehensive and purpose-built benchmark specifically for the evaluation of CMTs, taking inspiration from existing benchmarks while addressing the unique challenges of CMT evaluation.

\section{CUB: A Context Utilisation Benchmark}\label{sec:cub}

Given a CMT, \textsc{CUB} is designed to systematically test the technique across different datasets, models and metrics, providing rigorous evaluation. To ensure fair and meaningful comparisons, \textsc{CUB} incorporates a carefully designed pre-defined method for hyperparameter search, eliminating potential bias in method comparisons. 

\subsection{Language Models}\label{sec:models}
\textsc{CUB} evaluates the sensitivity of CMTs across a carefully selected suite of 11 LMs to analyse how effectiveness varies by model family, scale, instruction-tuning, and deployment paradigm.
\textbf{GPT2-XL \& Pythia 6.9B} \citep{gpt2,pythia} are included for compatibility and comparability with prior work on CMTs, as they remain frequently used models in model interpretability studies \citep{ortu-etal-2024-competition,jin-etal-2024-cutting}. 
\textbf{Qwen 2.5 (1.5B, 7B, and 32B)} \citep{qwen2.5} is chosen to systematically analyse the effects of scale. Both base and instruction-tuned variants are included for each size to capture the impact of instruction tuning on context utilisation. We also include the API-based models \textbf{Cohere Command A (111B),\footnote{\url{https://cohere.com/blog/command-a}}} specialised for RAG, as well as \textbf{GPT-4.1 mini} and \textbf{GPT-4.1},\footnote{\url{https://openai.com/index/gpt-4-1/}} which are popular baseline models.
See \Cref{app:model-motivation} for experiments and evidence used to guide the model selection.

\begin{table}[t]
    \centering
    \footnotesize
    \setlength{\tabcolsep}{2.7pt}
    \begin{tabular}{l c r r r r}
    \toprule
        Dataset & Split & \#samples & \%Gold & \%Conflict. & \%Irrel. \\
    \midrule
        CounterFact & \emph{dev} & 198 & 33.3 & 33.3 & 33.3 \\
        & \emph{test} & 2,499 & 33.3 & 33.3 & 33.3 \\
        NQ & \emph{dev} & 198 & 33.3 & 33.3 & 33.3 \\
        & \emph{test} & 4,945 & 33.4 & 33.1 & 33.4 \\
        DRUID & \emph{dev} & 198 & 33.3 & 33.3 & 33.3 \\
        & \emph{test} & 4,302 & 43.5 & 56.1 & 0.4 \\
    \bottomrule
    \end{tabular}
    \caption{Statistics of the datasets that form \textsc{CUB}. `Conflict.' and `Irrel.' denote conflicting and irrelevant contexts, respectively.}
    \label{tab:datasets}
\end{table}

\subsection{Datasets}
To systematically evaluate how CMTs respond to different types of contextual information, \textsc{CUB} evaluates each CMT across CounterFact, NQ and DRUID (see \Cref{tab:datasets}).
The inclusion of these datasets is based on three key criteria that ensure comprehensive coverage: (i) diversity in task difficulty, (ii) diversity in realistic versus synthesised RAG scenarios, and (iii) high utilisation in related work, ensuring our findings connect to the broader literature. CounterFact represents a controlled causal language modeling task with carefully synthesised counterfactual contexts designed to conflict with model memory. NQ represents a popular and more realistic setup focused on RAG for open-domain QA of substantially greater difficulty, with contexts sampled from Wikipedia. DRUID is a cutting-edge dataset representing another critical RAG task -- automated fact-checking -- which requires reasoning over naturally occurring claims and evidence from the internet. 

For each dataset, we curate samples representative of the three types of contexts that are fundamental to realistic RAG scenarios: 1) \textbf{gold} contexts that are relevant, 2) \textbf{conflicting} contexts that are relevant but always contradict the LM's memory, and 3) \textbf{irrelevant} contexts that provide no information to solve the given question \citep{fang-etal-2024-enhancing}. This approach ensures that \textsc{CUB} captures a wide spectrum of challenges to RAG in real-world deployments. \Cref{tab:datasets} summarises the dataset statistics, with context lengths detailed in \Cref{fig:context-length}.
More details on the data and the collection process can be found below and in \Cref{app:data-collection}.

\paragraph{CounterFact}
To construct a CounterFact dataset, we first identify base fact triplets (e.g., ``Barack Obama - was born in - Hawaii.'') from LAMA \citep{petroni-etal-2019-language} that have been memorised by Pythia 6.9B, following the approach by \citet{saynova2025factrecallheuristicspure}. We base the CounterFact dataset on facts memorised by Pythia 6.9B to obtain a set of triplets likely to have been memorised by all \textsc{CUB} models, since LMs have been found to memorise more facts as they grow in size \citep{saynova2025factrecallheuristicspure}. We confirm this in \Cref{app:data-collection}; all \textsc{CUB} LMs are found to have memorised at least 70\% of the underlying fact triplets.
These memorised fact triplets serve as the gold contexts.
Based on them, we also sample conflicting contexts (e.g., replacing ``Hawaii'' with ``France'') following the approach of \citet{meng2022locating}. 
For the irrelevant contexts, we randomly sample fact triplets unrelated to the target query. 

\paragraph{NQ}
The gold context samples are simply the original NQ samples. 
For the collection of samples with conflicting contexts, we use a substitution approach following \citet{longpre-etal-2021-entity}. 
For the collection of samples with irrelevant contexts, we apply a LM re-ranker to identify the most relevant non-gold paragraph from the Wikipedia page in which the gold context was found. With this approach, we collect irrelevant adversarial contexts representative of real-world RAG scenarios. 

\paragraph{DRUID}
The $\langle$claim, evidence$\rangle$ samples of DRUID have been manually annotated for stance of the evidence (supports, refutes, insufficient or irrelevant). We map stance to context type as described in \Cref{app:data-collection}. 
Since DRUID represents a reasoning task, asking the model whether provided evidence supports the claim under consideration (True or False), or is insufficient (None), the output space for the samples is limited to three tokens.

\subsection{Metrics}

Similarly to \citet{jin-etal-2024-cutting} we use a binary score to measure context utilisation. We refer to it as the \emph{binary context utilisation} ($\mathrm{BCU}$) score and define it as follows.
For relevant contexts (gold and conflicting) the score is 1 if the LM prediction is the same as the token promoted by the context, $t_C$, and 0 otherwise. For irrelevant contexts the score is 1 if the LM prediction is the same as the memory token, $t_M$, (i.e. the prediction made by the model before any context has been introduced) and 0 otherwise. 
We report the averaged $\mathrm{BCU}$ score per context type.
To assess the relative effectiveness of CMTs, we also report the net gain of each CMT, compared to when no CMT is applied, using $\mathrm{BCU}$ score ($\Delta=\mathrm{BCU_{\texttt{CMT}}}-\mathrm{BCU_{\texttt{Regular}}}$). We also consider \emph{continuous context utilisation}, $\mathrm{CCU}$, a more fine-grained metric that measures the change in outputted token probabilities as context is introduced, further described in \Cref{app:metrics}. Examples showcasing the behaviour of the BCU and CCU metrics can be found in the case-by-case-analysis in \Cref{app:results}.

\subsection{Hyperparameter Search} \label{sec:hpsearch}

For CMTs requiring hyperparameter tuning, we use the validation set of each dataset to select values that maximise the average BCU across all context types, unless a method-specific tuning procedure is explicitly specified (see \Cref{app:hyperparameter_search}).
This ensures a fair comparison between CMTs.

\section{Context Utilisation Manipulation Techniques}\label{sec:cmts}
\input{Tables/cmtsSummary}

We benchmark a comprehensive set of seven state-of-the-art CMTs on \textsc{CUB}, strategically selected to represent all major categories of CMTs and provide unprecedented coverage of the CMT landscape. 
Table \ref{tab:cmts_summary} provides a detailed overview of the key characteristics of each CMT, including their primary objectives, intervention levels, and computational costs for both tuning and inference. All CMTs that are benchmarked have been developed for singleton standard-length context settings. We adaptively select the appropriate set of LMs for each CMT based on compatibility. As a baseline for comparison, we also evaluate regular LMs on identical inputs with no CMT applied (\texttt{Regular}), enabling direct assessment of CMT effectiveness.

\paragraph{Fine-tuning}

We adapt the approach of \citet{li-etal-2023-large}, which fine-tunes LMs to ensure the usage of relevant contexts (see \Cref{app:finetuning}). It considers four different types of contexts: relevant, irrelevant, empty, and conflicting contexts.

\paragraph{Prompting}

We curate a set of 12 prompts for each evaluation dataset, tailoring prompt selection to each evaluated model's characteristics. Each prompt set combines human expertise and AI generation: 6 prompts are curated by human experts following established methodologies \citep{jin-etal-2024-cutting}, while 6 prompts are generated by advanced LLMs,\footnote{Mainly by ChatGPT, but also by Microsoft Co-pilot.} following established prompt engineering approaches \citep{wu2025axbenchsteeringllmssimple}.

\paragraph{Multi-agent}

Inspired by LM agents and self-refinement \citep{10.5555/3692070.3692537, feng-etal-2024-dont, madaan2023selfrefine}, which are widely adopted techniques in reasoning tasks, we decompose context utilisation into two components -- relevance and context faithfulness -- and assign each as a separate task to an individual LM agent (see \Cref{appendix:multiagent}).
We aim to examine whether LMs are capable of accurately evaluating context relevance and answer faithfulness, to subsequently self-correct themselves for improved faithfulness to relevant contexts.

\paragraph{Mechanistic interventions: PH3}

We adopt the PH3 method by \citet{jin-etal-2024-cutting} (see \Cref{app:ph3}). PH3 can be used in two different modes -- suppressing context attention heads (\texttt{PH3 +memory}) or suppressing memory attention heads (\texttt{PH3 +context}).

\paragraph{Context-aware contrastive decoding: ACD and COIECD}

Contrastive decoding approaches adjust the model's output distribution based on two distributions: one for which only the query is given as input and one for which the context also is included.
Among them, \emph{contextual information-entropy constraint decoding} (\texttt{COIECD}; \citealp{yuan-etal-2024-discerning}) is designed to detect the presence of knowledge conflicts and selectively resolve them, aiming to improve faithfulness to conflicting context without compromising performance when no conflict exists.
In contrast, \emph{adaptive contrastive decoding} (\texttt{ACD}; \citealp{kim-etal-2024-adaptive}) addresses the challenge of irrelevant context by using entropy-based weighting to adaptively ensemble parametric and contextual distributions. We test both on \textsc{CUB} to cover the nuance in decoding approaches.

\section{Features Impacting Context Utilisation}\label{sec:features}

To deepen our understanding of the results on \textsc{CUB}, we complement the benchmark with a sophisticated analysis of features likely to impact context utilisation. Our goal is to uncover the fundamental factors that determine \emph{why} certain CMTs and LMs achieve superior performance. We systematically investigate features at both model and input levels, providing unprecedented insights into the mechanisms underlying context utilisation.

\subsection{Model Features}\label{sec:model-features}

By virtue of the unprecedented LM coverage in \textsc{CUB}, we can systematically measure multiple salient model features. We analyse \textbf{model size}, whether the model is \textbf{instruction-tuned}, and \textbf{strength of model memory}. To control for external confounders related to model family and implementation differences, we carefully measure correlations with model size and instruction-tuning exclusively across Qwen models. Strength of model memory is quantified as the softmaxed logits for the top token predicted by the LM when only the query is provided (without context).

\begin{figure*}[ht]
    \centering
    \includegraphics[width=0.9\linewidth]{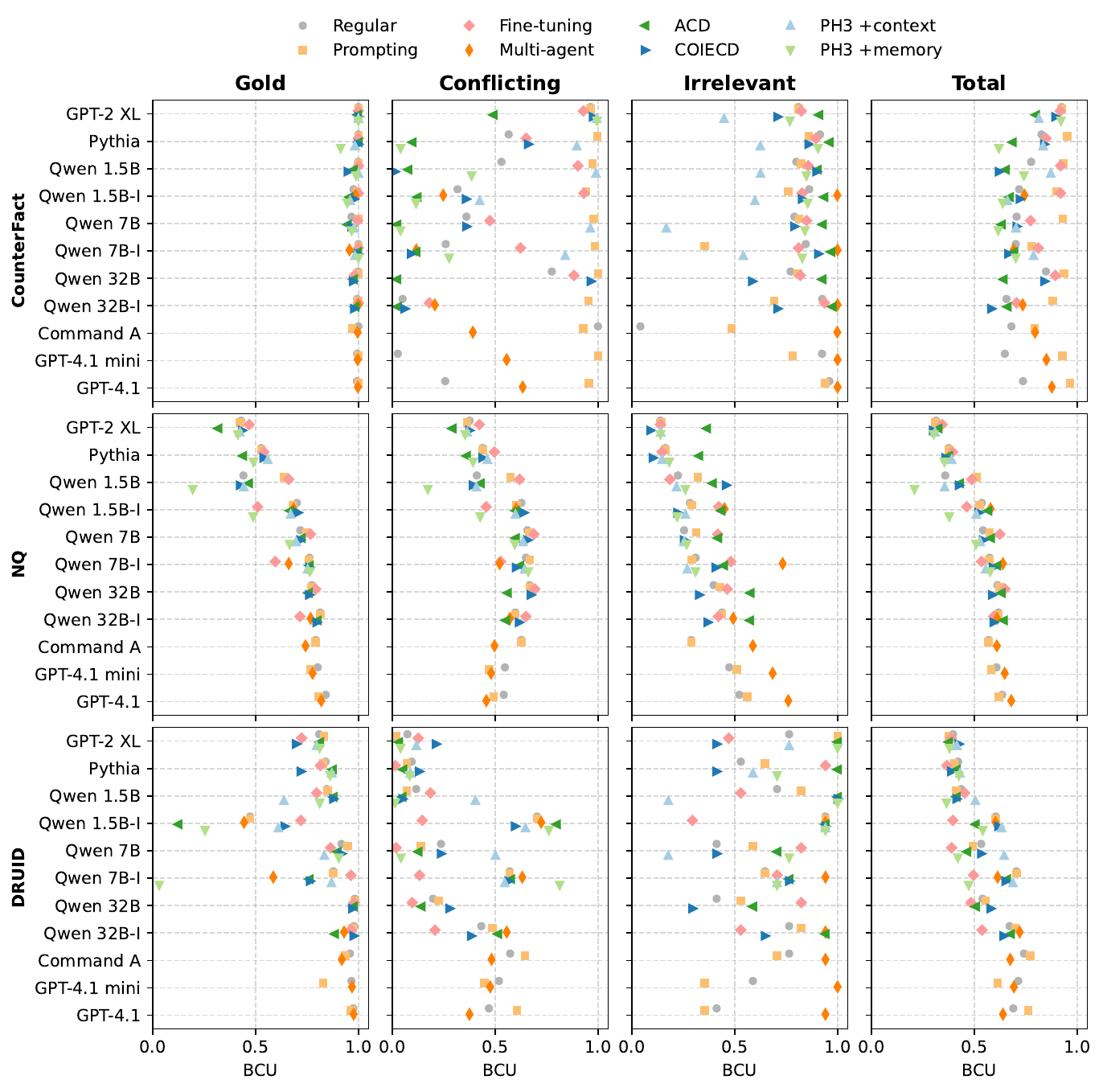}
    \caption{$\mathrm{BCU}$ scores for the evaluated context utilisation manipulation methods applied to the evaluated models and datasets. `Total' denotes the averaged performance across all context types.
    }
    \label{fig:BCU}
\end{figure*}
\begin{figure}
\scriptsize
\centering
\includegraphics[width=\linewidth]{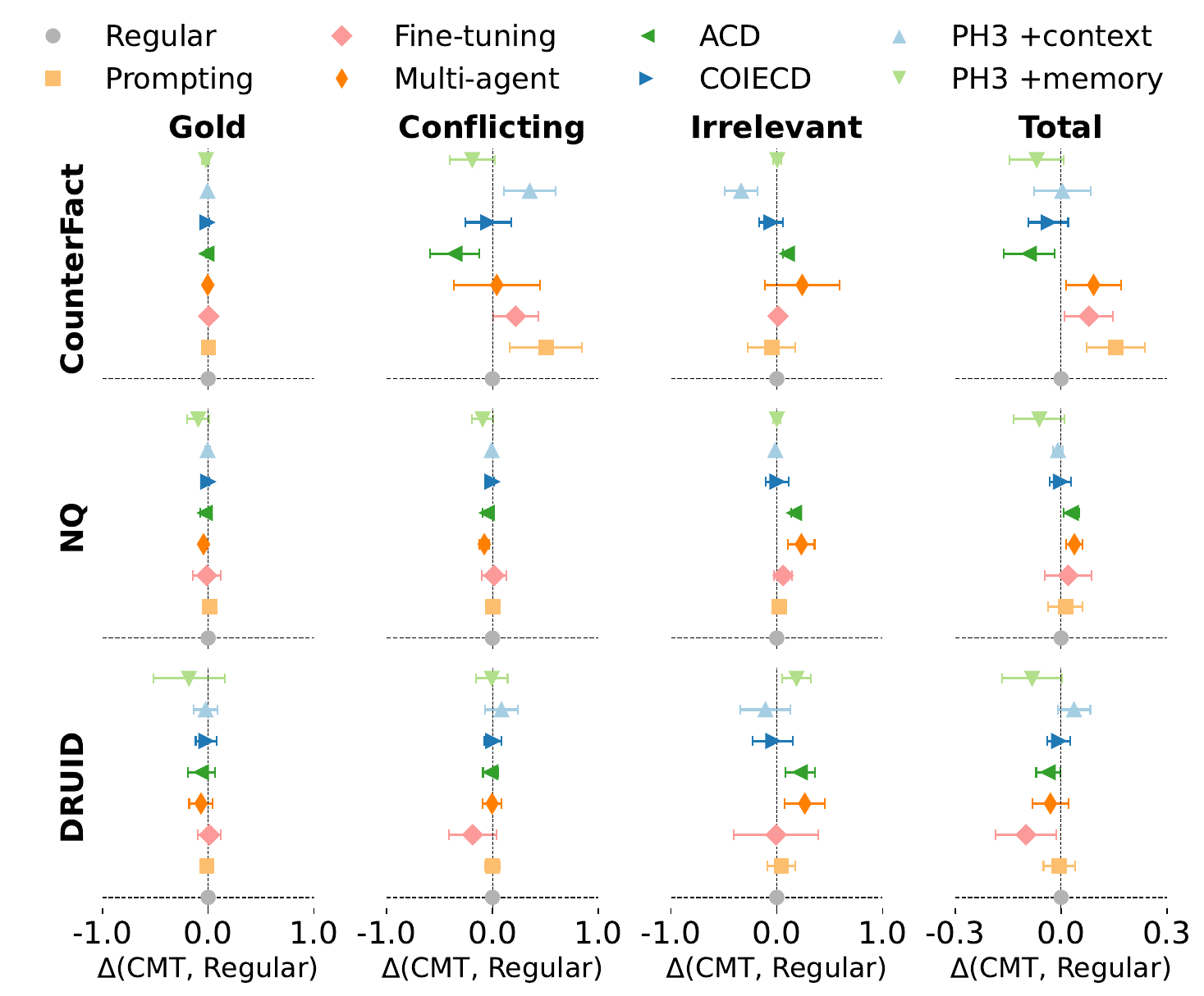}
\caption{Model-averaged relative performance ($\Delta$) of each CMT compared to \texttt{Regular} across datasets and context types. The horizontal bars represent the standard deviation.}
\label{fig:reg_cmt_diff_avg}
\end{figure}

\subsection{Input Features}\label{sec:input-features}

We measure multiple input characteristics found to impact context utilisation for humans and/or LMs (see \Cref{app:features}). By considering \textbf{context length} and \textbf{Flesch reading ease score}, we aim to measure whether the context is \emph{difficult to understand} \citep{gao2024retrievalaugmentedgenerationlargelanguage,vladika-matthes-2023-scientific}. Using \textbf{distractor rate}, we aim to measure whether the context contains \emph{distracting information} \citep{shaier-etal-2024-desiderata}. With \textbf{query-context overlap} we also aim to measure \emph{query-context similarity} \citep{wan-etal-2024-evidence}. Lastly, we check the \textbf{answer position} \citep{liu-etal-2024-lost} and if the evaluated LMs find the context \textbf{relevant}.

\subsection{Metric for Feature Impact}

By virtue of the unified and controlled setup of \textsc{CUB}, we can systematically study correlation coefficients to investigate the impact of different input and model features with minimal risk of confounding factors. We employ Spearman's $\rho$ to measure the impact of features on context utilisation, proxied by BCU, providing robust statistical insights into the factors that drive CMT effectiveness. 

\section{Main Results on CUB} \label{sec:results}

The \textsc{CUB} results can be found in \Cref{fig:BCU,fig:reg_cmt_diff_avg}; $\mathrm{CCU}$ scores together with additional results and analyses, qualitative as well as quantitative, can be found in \Cref{app:results}. We structure our analysis around a set of key findings about CMT effectiveness.

\subsection{Overall Trends}\label{sec:trends}
We first note that the $\mathrm{BCU}$ and $\mathrm{CCU}$ scores in \Cref{fig:BCU,fig:CCU}, respectively, support the same trends and focus the analysis on the $\mathrm{BCU}$ results.

\paragraph{Context utilisation scales with model size -- but with surprising exceptions.} 
From \Cref{fig:BCU}, we observe that larger \texttt{Regular} LMs generally achieve superior performance across all context types for NQ and DRUID, following expected scaling laws.
On NQ, the best performing \texttt{Regular} model is GPT-4.1, and on DRUID it is Command A. Remarkably, our results reveal that applying the right CMT to a smaller LM can achieve context utilisation performance on par with much larger regular LMs -- for instance, \texttt{Fine-tuning} applied to Qwen 7B matches the performance of \texttt{Regular} Qwen 32B on NQ. 
However, CounterFact reveals a striking counter-intuitive pattern: \texttt{Regular} model performance across all contexts generally \emph{decreases} as model size \emph{increases}. 
We hypothesise this unexpected phenomenon stems from two related factors.
First, the atomic and sentence-level contexts in CounterFact contain relatively less rich contextual information compared to real-world contexts from NQ and DRUID.
Second, larger LMs often exhibit a stronger belief in their parametric knowledge, making them more `stubborn' when presented with conflicting information \citep{xie2024knowledgeconflict}.
Consequently, the sparse context from CounterFact is often insufficient to overwrite the more deeply ingrained knowledge of larger models, leading to the observed performance decline.

\paragraph{Most CMTs exhibit inflated performance on conflicting CounterFact contexts.} We observe a striking phenomenon: all LMs that do not already achieve perfect BCU scores on conflicting CounterFact contexts leap to a perfect score near 1.0 under \texttt{Prompting}, \texttt{PH3 +context}, and \texttt{Fine-tuning}. 
However, these same CMTs fail to deliver comparable improvements on the more realistic NQ or DRUID datasets. These results expose a critical flaw in current evaluation practices -- CMTs that appear highly effective in simplified settings may fail to generalise to real-world complexity. This finding demonstrates the essential need for holistic evaluation frameworks like \textsc{CUB}.

\subsection{CMT Comparison}

We further assess whether the CMTs consistently outperform \texttt{Regular} across different context types.
Figure \ref{fig:reg_cmt_diff_avg} shows the average $\Delta$ of each CMT, aggregated over all evaluated models.
A value above zero indicates that the CMT yields a net improvement over \texttt{Regular}, whereas a negative value highlights cases where the CMT degrades performance.

\paragraph{We uncover a fundamental trade-off between sensitivity to relevant contexts versus robustness to irrelevant contexts.} 
Our analysis reveals a striking pattern: each CMT exhibits inherent trade-offs across context types, with overall $\Delta$ values (Total) converging to near zero across NQ and DRUID. This convergence exposes that \emph{no single CMT emerges as universally superior}.
For instance, \texttt{PH3 +context} demonstrates consistent improvements over \texttt{Regular} in conflicting contexts, but significantly underperforms when applied to irrelevant contexts.
Conversely, \texttt{ACD}, while effectively handling irrelevant contexts, shows degraded performance in conflicting context scenarios. 

In realistic RAG scenarios, the type of context provided to the LM cannot be predicted in advance. Therefore, the ideal CMT must perform optimally across \emph{all} context types. Our comprehensive evaluation reveals that while we have identified CMTs that excel with either relevant or irrelevant contexts in isolation, \emph{no existing CMT can robustly handle both relevant and irrelevant contexts simultaneously}. This represents a fundamental gap in current CMT capabilities that needs to be addressed.

Meanwhile, prompting-based CMTs, such as \texttt{Prompting} and \texttt{Multi-agent}, demonstrate remarkably stable performance across context types, avoiding the dramatic drops observed with other CMTs.
Compared to more complex CMTs, they offer this robustness with lower optimisation and implementation costs, making them attractive for practical deployment.
\texttt{Multi-agent} shows particularly promising results with clear gains in irrelevant contexts, though its efficacy remains limited in gold and conflicting settings.
This pattern suggests that while LMs demonstrate competence in identifying irrelevant contexts, they face fundamental limitations in effectively utilising relevant information.

\subsection{Trade-off Analysis via Pareto Frontiers}
\input{Tables/pareto}

Building upon the general trends established in Section~\ref{sec:trends}, we analyse the fundamental trade-off between faithfulness ($Avg(\mathrm{BCU_{\texttt{Gold}}}, \mathrm{BCU_{\texttt{Conflicting}}})$) and robustness ($\mathrm{BCU_{\texttt{Irrelevant}}}$) through Pareto frontier analysis across various LMs and CMTs.
This analysis enables the identification of (LM, CMT) configurations that achieve an optimal balance between the two metrics, specifically highlighting pairs where performance in one dimension cannot be improved without compromising the other.

As illustrated in Table~\ref{tab:pareto}, optimal performance is highly sensitive to both dataset characteristics and model capacity.
Specifically, prompting with large-scale LMs (e.g., Qwen 32B) consistently defines the high faithfulness region of the frontier.
Conversely, the \texttt{Multi-agent} approach emerges as the most robust CMT configuration, frequently anchoring the high-robustness end of the spectrum across all evaluated datasets.
It is observed, however, that the efficacy of the \texttt{Multi-agent} approach is intrinsically linked to the underlying LM's reasoning and instruction-following capabilities.

Intermediate trade-offs are often provided by \texttt{fine-tuning} and decoding approaches (\texttt{ACD} and \texttt{COIECD}), which offer a balance between the two metrics depending on the specific dataset.
These results confirm that no single CMT universally dominates across all datasets and metrics. 
Instead, the selection of an optimal (LM, CMT) pair must be guided by the specific faithfulness-robustness priorities dictated by the target application. 

\subsection{Impact of Model and Input Features}

\input{Tables/featureCorrs/model_agg}
\input{Tables/featureCorrs/input_agg}

\Cref{tab:corr_model,tab:corr_input} present Spearman's $\rho$ between BCU and the features described in \Cref{sec:features}.

\paragraph{Larger LMs perform better on NQ and DRUID.} Corroborating our findings in \Cref{sec:trends}, we observe a positive correlation with model size ($\rho \approx 0.3$) on DRUID gold contexts.
\texttt{Multi-agent} also works significantly better with bigger LMs on DRUID gold contexts ($\rho = 0.42$). 
In addition, we observe a positive correlation with model size on NQ gold contexts ($\rho \in [0.20, 0.37]$).

\paragraph{Instruction-tuning is beneficial for conflicting and irrelevant DRUID contexts.} We note how instruction tuning generally correlates with improved performance on conflicting and irrelevant DRUID contexts ($\rho \in [0.29, 0.77]$). The conflicting DRUID contexts frequently require the LM to abstain (i.e. respond with a `None') when presented with insufficient contexts -- instruction-tuned models are likely more adept at this. 

Conversely, instruction-tuning is clearly detrimental for conflicting CounterFact contexts ($\rho \leq -0.36$), potentially because this makes the LMs more critical of unreliable information, as opposed to performing pure causal language modelling.

\paragraph{A strong model memory corresponds to high performance on irrelevant contexts from NQ and CounterFact.} 

We observe high correlations ($\rho \approx 0.36$) between memory strength and robustness to irrelevant contexts for \texttt{Regular} on CounterFact and NQ. These correlations increase when \texttt{Fine-tuning}, \texttt{ACD} or \texttt{Prompting} is applied. Furthermore, we observe for CounterFact how strong \texttt{Regular} model memory correlates with low performance on conflicting contexts ($\rho = -0.44$). This is also supported by previous works, which have shown how LMs are resistant to synthesised contexts that contradict the internal model memory \citep{longpre-etal-2021-entity,xie2024knowledgeconflict}.

\paragraph{Answer position matters little for context utilisation.} We measure low correlation values (below 0.3) across all settings for answer position in the context and Flesch reading ease score, and have thus omitted them in \Cref{tab:corr_input}. Previous work has already found the Flesch reading ease score to show low correlations with LM context utilisation; our work further supports this finding \citep{hagstrom-etal-2025-reality}. \citet{liu-etal-2024-lost} found the answer position impactful for the utilisation of long contexts. \textsc{CUB} does not contain equally long contexts, which potentially explains why we do not see the same impact of answer position.

\paragraph{Context utilisation on gold NQ contexts is degraded on long contexts with high distractor rates.} We measure weak negative correlations with context length ($\rho = -0.20$) and distractor rate ($\rho = -0.15$) with respect to \texttt{Regular} performance on gold NQ contexts. This is expected -- long gold contexts or contexts with a high rate of distractors should be more difficult to process and utilise. We hypothesise the fairly low correlation levels are a consequence of each feature alone not being sufficiently predictive of model performance.

\section{Conclusion}

We present \textsc{CUB}, the first comprehensive benchmark for systematically evaluating CMTs across diverse contexts, datasets, and model architectures. 
Using \textsc{CUB}, we conduct the most extensive comparison to date of seven representative CMTs, revealing a core trade-off between robustness to irrelevant context and effective use of relevant information. Our analysis shows that model characteristics play a dominant role in context utilisation, while input features alone are weak predictors. Crucially, we uncover major flaws in current evaluation practices -- many CMTs perform well on synthetic datasets but falter in more realistic settings. These findings underscore the need for holistic evaluation frameworks and set a new standard for developing CMTs in real-world RAG environments.

\section*{Limitations}

While context can be characterised by various dimensions, including its length or the number of documents provided, our current study specifically focuses on context type.
We view the investigation of longer or multi-document contexts as important fields for further research, yet orthogonal to our current focus, as explained below.

\emph{Long-context} settings (e.g., over 4k tokens) have recently garnered significant attention \citep{liu-etal-2024-lost,mujahid2025stresstestingfactualconsistency}, and evaluating CMTs in these scenarios is an important direction for future research. However, this paper focuses on a different set of challenges -- those associated with standard-length contexts -- making long-context evaluation out of scope. Long-context tasks involve distinct issues in context utilisation and require different evaluation strategies and CMTs \citep{shaham-etal-2023-zeroscrolls, zhang-etal-2024-fine, min-etal-2023-factscore, zhang2024middlelanguagemodelsuse}. Our goal with \textsc{CUB} was to introduce a foundational benchmark that enables rigorous comparisons between CMTs developed for standard-length contexts across diverse types of contexts and domains. Starting with standard-length contexts was a natural choice, as they remain both prevalent and crucial -- standard-length contexts are featured in many popular datasets for studies of context utilisation, such as CounterFact, NQ, and DRUID. Even within this scope, \textsc{CUB} uncovers valuable insights into CMT behavior. Future work can extend on the foundation provided by \textsc{CUB} to include long-context benchmarks.

The \emph{multi-document} setting -- where multiple contexts are provided to the model -- is another important area for future research \cite{sun2025evaluationframeworkhighlightexplanations}. However, \textsc{CUB} focuses on single-document settings to allow controlled and meaningful evaluation of diverse CMTs across different context types. Evaluating CMTs in a multi-document setup introduces significant complexity, as it becomes challenging to disentangle the influence of individual documents from their combined effects. Furthermore, many of the recent and widely used CMTs included in \textsc{CUB} were designed specifically for single-document scenarios and are not yet compatible with multi-document inputs. As such, \textsc{CUB} provides a necessary and robust foundation, which future work can build upon to address the more intricate challenges of multi-document settings.

Another limitation of \textsc{CUB} is its \emph{coverage of only a subset of available models}. However, the thoughtful model selection for \textsc{CUB} ensures that the insights it provides are broadly informative and likely to generalise -- \textsc{CUB} is built on a carefully curated selection of 11 models, chosen to balance recency, diversity in model size, instruction tuning, and relevance to RAG-specific applications. Reasoning models would also have been interesting to study but were not included to preserve compatibility with the analysed CMTs and to facilitate fair comparisons between the models in the benchmark. Furthermore, the \texttt{Multi-agent} already covers a more reasoning-focused setting. Additionally, \textsc{CUB} is designed to be extensible, allowing future inclusion of a wider range of models.

While the CMTs included in \textsc{CUB} represent the current state of the art and cover all major categories, the rapidly evolving landscape means that \emph{many promising methods remain to be benchmarked}. However, by open-sourcing \textsc{CUB}, we provide the research community with a robust foundation for systematic evaluation of both existing and future CMTs.

The datasets used in \textsc{CUB} were carefully selected to span a range of task difficulties and reflect common RAG scenarios. However, the \emph{insights derived from \textsc{CUB} are inherently constrained by the characteristics of these underlying datasets}. Additionally, since all datasets are in English, it remains an open question whether the findings generalise to other languages \citep{chirkova-etal-2024-retrieval}. Future work could extend \textsc{CUB} to cover more datasets as well as non-English datasets.

Finally, \textsc{CUB} does not feature datasets with naturally occurring \emph{temporal dynamics}, which represent an interesting avenue for expansions of the benchmark. Time-sensitive information can introduce natural conflicts within the context, offering a richer setting to analyse context utilisation \citep{loureiro-etal-2022-tempowic, xiong-etal-2024-large,marjanovic-etal-2024-dynamicqa}.
Instead, \textsc{CUB} leverages synthesised conflicts to mimic temporal changes, offering a controlled environment that is particularly useful for evaluating models with different knowledge cut-off dates and consistent with common practice in context utilisation research. Moreover, while the current context types in \textsc{CUB} were carefully chosen based on prior work on RAG \citep{fang-etal-2024-enhancing}, other potentially valuable context types may have been overlooked.

\section*{Acknowledgments}
$\begin{array}{l}\includegraphics[width=1cm]{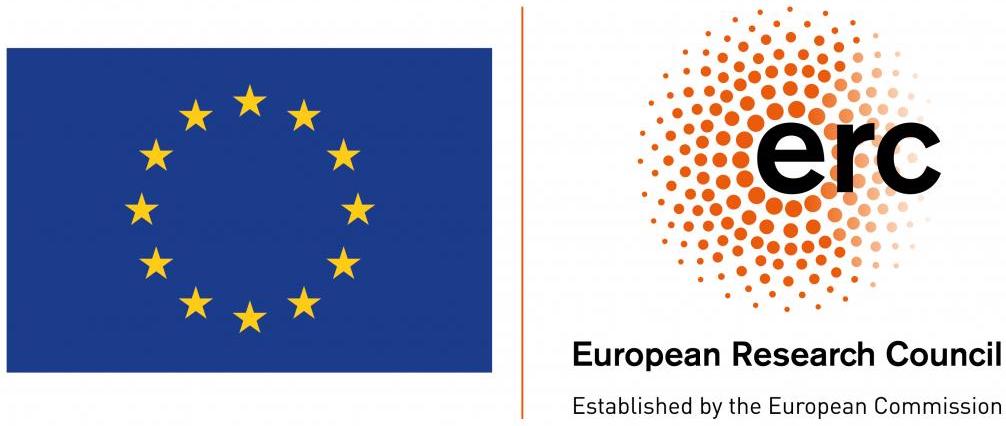} \end{array}$ 
This research was co-funded by the European Union (ERC, ExplainYourself, 101077481), by the Pioneer Centre for AI, DNRF grant number P1, as well as by The Villum Synergy Programme. Views and opinions expressed are however those of the author(s) only and do not necessarily reflect those of the European Union or the European Research Council. Neither the European Union nor the granting authority can be held responsible for them.

This research was also co-funded by the Wallenberg AI, Autonomous Systems and Software Program (WASP) funded by the Knut and Alice Wallenberg Foundation. The computations were enabled by resources provided by the National Academic Infrastructure for
Supercomputing in Sweden (NAISS) at Alvis partially funded by the Swedish Research Council through grant agreement no. 2022-06725.

This work was partially supported by the National Research Foundation of Korea(NRF) grant funded by the Korea government(MSIT and MOE: RS-2025-16063688, MSIT: RS-2025-02215813, RS-2026-25491306, and RS-2025-00562784) and Institute of Information \& communications Technology Planning \& Evaluation (IITP) grant funded by the Korea government (MSIT, RS-2021-II211343)

Lastly, we thank the anonymous reviewers for their insightful comments and helpful feedback.

\bibliography{anthology,custom}

\appendix

\section{Additional results}\label{app:results}

\subsection{CUB results}
The exact BCU scores on \textsc{CUB} can be found in \Cref{tab:main_results_bcu}. We also report the accuracy scores on \textsc{CUB} in \Cref{tab:main_results_acc}. For CounterFact and DRUID, accuracy is measured based on whether the first generated token is the same as the first gold token. For NQ, for which the correct answer may be different permutations of the same set of tokens, we measure accuracy based on whether the first output token (e.g. ``July'') matches any of the tokens in the answer (e.g. ``15 July''). 

\begin{figure*}[ht]
    \centering
    \includegraphics[width=0.9\linewidth]{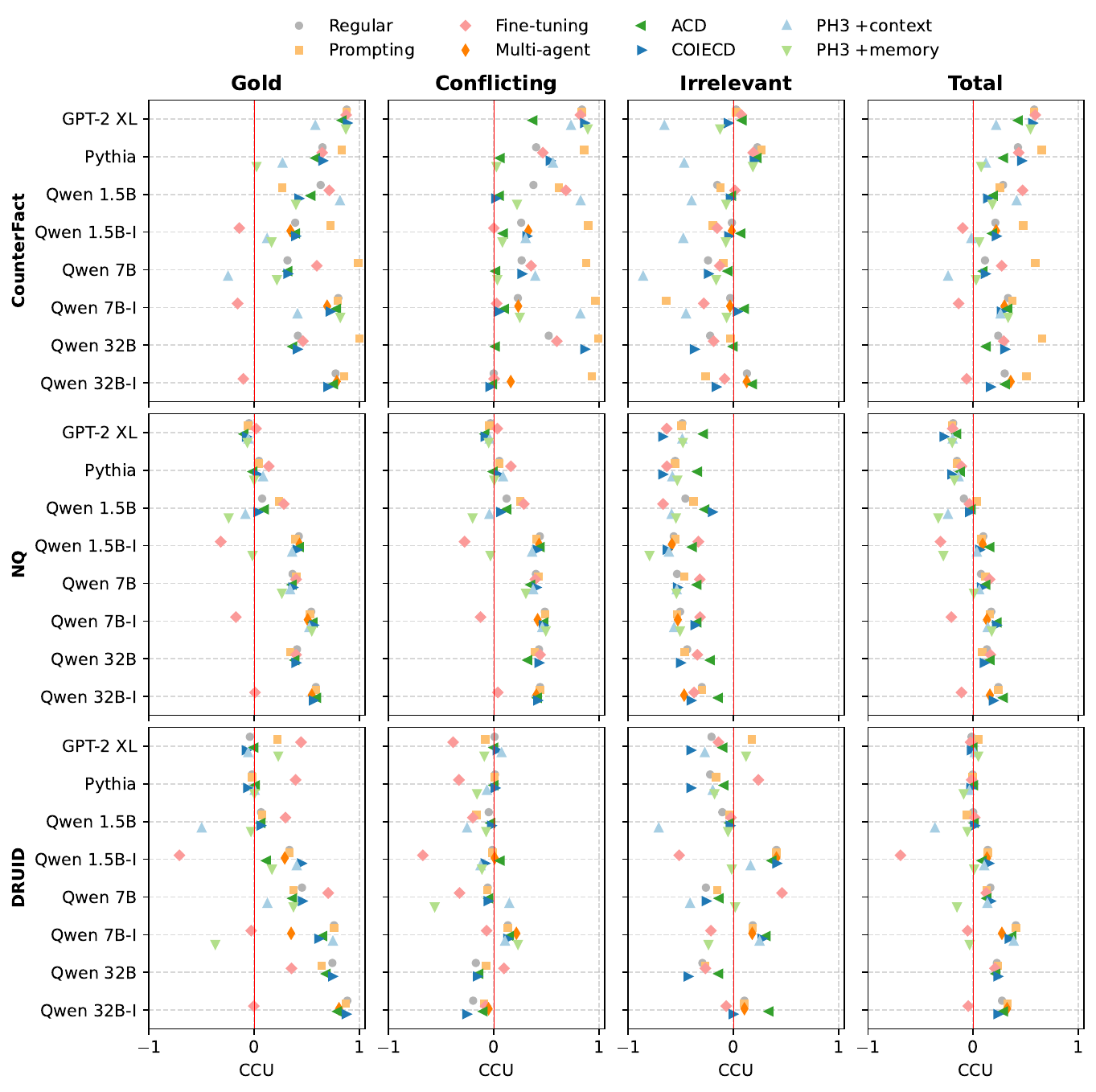}
    \caption{$\mathrm{CCU}$ scores for the evaluated context utilisation manipulation methods applied to the evaluated models and datasets. `Total' denotes the averaged performance across all context types. A high $\mathrm{CCU}$ score is desirable regardless of context type. The red vertical lines indicate scores of 0.}
    \label{fig:CCU}
\end{figure*}

$\mathrm{CCU}$ scores can be found in \Cref{fig:CCU}. For the $\mathrm{CCU}$ scores, we note that they generally follow the same trends as the $\mathrm{BCU}$ scores in \Cref{fig:BCU}; some CMTs perform better on gold, conflicting or irrelevant contexts, while none are superior when all context types are taken into consideration. The only disparate trend at odds with the $\mathrm{BCU}$ scores is that \texttt{Fine-tuning} Qwen models that have been instruction-tuned stand out by performing extra poorly with respect to $\mathrm{CCU}$ score. We hypothesise that this is a consequence of an increase in $P_M(t_C|Q)$ (i.e. prediction probability without context) from the fine-tuning, yielding less room for improvement in prediction confidence when context is introduced.

\input{Tables/mainTableBCU}
\input{Tables/mainTableAcc}

\subsection{Analysis of inflated CMT performance on CounterFact}

The inflated performance on CounterFact, observed in \Cref{fig:BCU,fig:reg_cmt_diff_avg}, can to a certain extent be explained by a suboptimal default prompt for CounterFact. Following previous work, the default prompt only contained the example to be completed, without any additional instructions or few-shot examples. For NQ and DRUID, the default prompt contained task instructions and few-shot examples. Furthermore, we observe how \texttt{Prompting} performs best on CounterFact on average, with a near perfect performance, indicating that a better default prompt may have neutralised any additional improvements from other CMTs. This raises the question of whether certain CMTs only address low context utilisation when caused by poor prompting, finding no leverage if the prompt already is adequate.

\subsection{Quality check of irrelevant NQ contexts}

For the \textsc{CUB} evaluation, we find 244 (14\%) NQ samples with the context type `irrelevant' for which at least 5 of the 9 evaluated LMs switch prediction to the gold answer \emph{after} having seen the sample context. This indicates that some of the irrelevant contexts may actually be gold, as a result of quality issues with the annotation for NQ (in our sampling we assume that Wikipedia paragraphs not annotated as gold are not gold). However, we also note for some of these 244 samples that the context may simply be the heading of a Wikipedia page with the same title as the gold answer (e.g. ``<H1> Scythe </H1>'' when the gold answer is ``scythe'' for the query ``what is the name of the weapon the grim reaper carries?''), without providing sufficient evidence with respect to the question, raising the question of whether they should be considered relevant by the model.

\subsection{Performance of Relevance Judgement}
\input{Tables/multiAgentRelevance}
For the \texttt{Multi-agent} technique, we investigate whether instruction-tuned LMs are capable of identifying irrelevant context when explicitly prompted to do so.
According to \Cref{tab:multi_agent_rel_acc}, the \texttt{Multi-agent} approach demonstrates strong performance in detecting irrelevant contexts and in recognising gold contexts as relevant.
Although it does not reliably maintain a closed-book response when directly generating responses (i.e. \texttt{Regular}), it can accurately detect irrelevance when equipped with an explicit relevance assessment setup.

The prediction accuracy of relevance assessment on conflicting contexts is consistently lower than that on other contexts.
This discrepancy is particularly evident in the conflicting contexts of the CounterFact dataset.
For instance, we found that LMs often generate feedback such as: ``X is Y, not Z. Therefore, the context is irrelevant''.
This suggests that LM interprets factual inconsistency with its internal knowledge as a signal of irrelevance, even when instructed to ignore its own memory.

One possible explanation for this behaviour lies in the nature of the CounterFact dataset itself.
Contexts in CounterFact are typically composed of single-sentence facts, which may lack sufficient surrounding information to render the context trustworthy from the model's perspective.
Such behaviour is less pronounced in NQ and DRUID datasets, where the provided contexts are relatively longer and richer, offering more semantic cues that may help the LM interpret the information as contextually anchored \citep{xie2024adaptive}.

The performance of relevance assessment is particularly low on the NQ dataset compared to other datasets.
Since irrelevant contexts of NQ dataset are sampled from the same document and may be topically or semantically similar to the question, distinguishing relevance may become more challenging.

\subsection{Case-by-case analysis}\label{app:case-by-case}

This section carries out a case-by-case analysis for the Qwen 7B model on \textsc{CUB} across gold, conflicting and irrelevant context types for CounterFact, NQ and DRUID. Using this analysis, we get a better qualitative understanding of how different CMTs perform on \textsc{CUB}.

\subsubsection*{CounterFact}

The case analysis for CounterFact is shown in \Cref{tab:case-counterfact-gold,tab:case-counterfact-conflicting,tab:case-counterfact-irrelevant}. Based on the sampled results below, we can see how the CMTs essentially perform the same for the gold context, while the CCU for the \texttt{Fine-tuning} method is slightly lower. 

For the conflicting context, we note that all CMTs show a similar performance, except for the \texttt{COIECD} and \texttt{Fine-tuning} methods. Under the \texttt{COIECD} method, the LM does not follow the context and makes a prediction (``Michigan'') that is not even aligned with the memory of the model (``Detroit''). Under the \texttt{Fine-tuning} method, the LM follows the context but assigns a lower probability to its prediction. 

For the irrelevant context, a wider spread in CMT performance can be observed. Only under the \texttt{Multi-agent} and \texttt{Fine-tuning} methods is the LM robust to the noisy context and sticks to its original prediction.

\input{Tables/case_by_case/counterfact}

\subsubsection*{NQ}

The case analysis for NQ is shown in \Cref{tab:case-nq-gold,tab:case-nq-conflicting,tab:case-nq-irrelevant}. We observe trends similar to those for CounterFact. One difference here is that the \texttt{Multi-agent} method is the only method that makes the LM robust to the irrelevant context.

\input{Tables/case_by_case/nq}

\subsubsection*{DRUID}

The case analysis for DRUID is shown in \Cref{tab:case-druid-gold,tab:case-druid-conflicting,tab:case-druid-irrelevant}. We note how the \texttt{PH3 +memory} performs particularly poorly for the gold context, and that the \texttt{Multi-agent} approach best adapts to conflicting contexts. Similarly, for the irrelevant context case we note how the \texttt{Fine-tuning} and \texttt{Multi-agent} approaches make the LLM more robust.

\input{Tables/case_by_case/druid}

\section{Model Selection}
\label{app:model-motivation}

This section provides additional details and experiments used to carefully guide the model selection for \textsc{CUB}. The model codes for the API based models are also provided here since the API providers may update the models to new versions in the future.

\paragraph{GPT-2 XL and Pythia 6.9B} As already described in \Cref{sec:models}, GPT-2 XL and Pythia 6.9B were included to preserve compatibility with prior work on CMTs. GPT-2 XL is a lightweight model that is inherently easier to modify and interpret compared to larger models. As for Pythia 6.9B, the Pythia model suite was explicitly designed to enable research in interpretability.

\paragraph{Qwen} Models from the Qwen2.5 model suite \citep{qwen2.5} were included to represent a relevant, recent and performant set of models for which we can experiment with different model sizes and instruction-tuning.

The inclusion of Llama 3.1 was also considered in an initial analysis. However, since initial results showed how the Qwen model performed on par with the Llama model, and since the Qwen models cover a broader range of model sizes with and without instruction-tuning, we opted to only include Qwen instead of Llama to yield more generalising insights. Llama 3.1 only corresponds to three model sizes (8B, 70B, 405B), of which the two larger model sizes are too great for most academic compute resources. 

\begin{figure}
    \centering
    \includegraphics[width=0.75\linewidth]{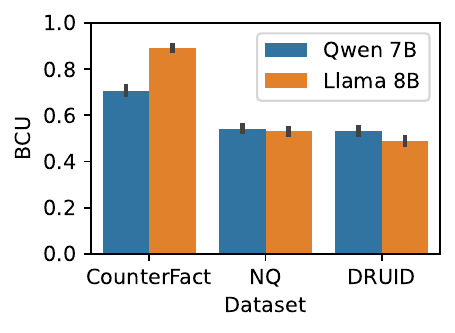}
    \caption{Averaged BCU scores across all context types for Qwen and Llama on the \textsc{CUB} datasets.}
    \label{fig:qwen-llama-tot}
\end{figure}

\begin{figure*}[h]
\centering
\begin{subfigure}[t]{0.33\textwidth}
\includegraphics[width=\linewidth]{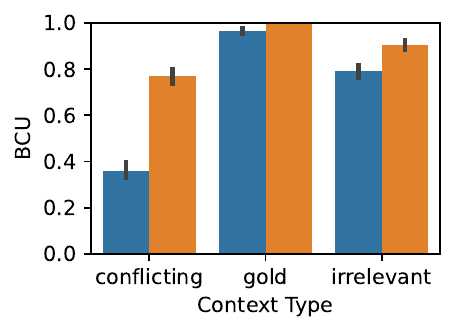} 
\caption{CounterFact}
\label{fig:qwen-llama-counterfact}
\end{subfigure}%
\begin{subfigure}[t]{0.33\textwidth}
\includegraphics[width=\linewidth]{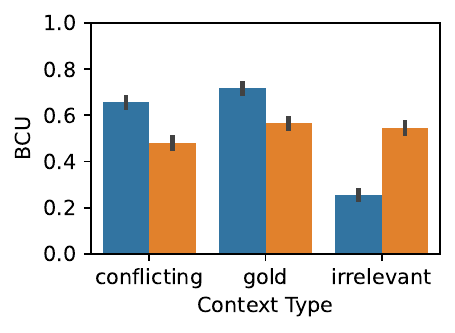}
\caption{NQ}
\label{fig:qwen-llama-nq}
\end{subfigure}%
\begin{subfigure}[t]{0.33\textwidth}
\includegraphics[width=\linewidth]{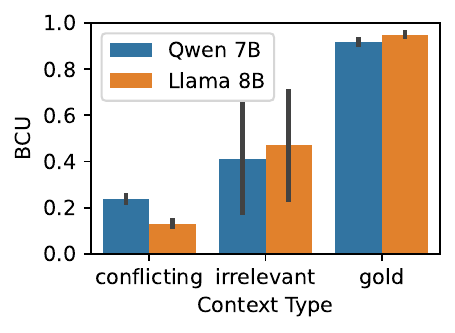}
\caption{DRUID}
\label{fig:qwen-llama-druid}
\end{subfigure}
\caption{BCU scores for Qwen and Llama on the \textsc{CUB} datasets.}
\label{fig:qwen-llama}
\end{figure*}

Regular mode Llama 8B\footnote{\texttt{meta-llama/Llama-3.1-8B} from Hugging Face.} is compared to Qwen 7B in \Cref{fig:qwen-llama-tot,fig:qwen-llama}. As can be seen from the results, Qwen and Llama have a similar performance on CounterFact, NQ and DRUID. Llama outperforms Qwen on CounterFact, while Qwen slightly outperforms Llama on NQ and DRUID. Based on these results, we conclude that Qwen 7B should work as a good representative also for the Llama 8B model on \textsc{CUB}.


The Gemma 3 models\footnote{E.g. \texttt{google/gemma-3-12b-pt} from Hugging Face.} were also considered but not included in \textsc{CUB}. Like the Qwen model family, the Gemma model family also covers various model sizes and instruction tuning. However, based on our initial analysis we concluded that incorporating Gemma into CUB was unlikely to yield any significant new findings due to Qwen already being included: Both Qwen and Gemma support the same context window size and show comparable performance in various NLP benchmarks.\footnote{Statistics reported by \url{llm-stats.com} which compares Gemma 3 12B and Qwen2.5 14B (the most comparable model sizes) show that both models have a comparable performance on many benchmarks, such as HumanEval and GPQA.} 

\paragraph{API model codes} The code for the Cohere Command A model is \texttt{command-a-03-2025}. The codes for the GPT-4.1 models from OpenAI are \texttt{gpt-4.1-2025-04-14} and \texttt{gpt-4.1-mini-2025-04-14}. These model codes correspond to the most recent model version available at the time of writing.

\section{Data Collection}
\label{app:data-collection}

We collect three datasets for \textsc{CUB}: CounterFact, NQ, and DRUID. For each dataset, we carefully construct validation and test splits to enable fair and unified hyperparameter tuning for each CMT, as well as unbiased final evaluation metrics. The following subsections describe the construction and characteristics of each dataset in detail.

\subsection{CounterFact}
Samples from the CounterFact dataset can be found in \Cref{tab:counterfact-examples}. The relations covered by the dataset are \emph{capital of} (80\%), \emph{country of origin} (9\%), \emph{location of formation} (9\%), \emph{field of work} (1\%) and \emph{country of citizenship} (1\%).

\begin{table}[!t]
    \centering
    \scriptsize
    \begin{tabular}{p{2.1cm} p{2.1cm} r l}
    \toprule
    Prompt & Context & \#char. & Type \\
    \midrule
        Athens, the capital city of & Fact: Athens, the capital city of Greece. & 41 & Gold \\
    \midrule
        Thomas Ong is a citizen of & Fact: Thomas Ong is a citizen of Pakistan. & 42 & Conflicting \\
    \midrule
        Prince Oscar Bernadotte is a citizen of & Fact: Melbourne, that is the capital of Jordan. & 47 & Irrelevant \\
    \bottomrule
    \end{tabular}
    \caption{CounterFact samples and corresponding context types. `\#char.' indicates the length of the context, measured by number of characters.}
    \label{tab:counterfact-examples}
\end{table}

\paragraph{Rate of memorisation of \textsc{CUB} models}
We evaluate all \texttt{Regular} LMs on the samples from \textsc{CUB} CounterFact without context. The results can be found in \Cref{tab:counterfact-gold-memorisation}. We observe rates above 70\% for all models. As expected, the highest memorisation rate is found for Pythia. The lowest is found for GPT-2 XL, which also can be expected as the model is quite small.

\input{Tables/counterfact_memorisation.tex}

\paragraph{Prompt templates}
Following the same approach as previous work, no specific prompt template was used for the LMs evaluated on CounterFact. The LMs were evaluated in a simple sentence completion format as shown in \Cref{tab:counterfact-examples}.

However, since the sentence completion format is less compatible with the instruction-tuned models, we added a small prompt template for the evaluation of the instruction-tuned Qwen models on CounterFact, as follows.
\begin{lstlisting}[frame=single, title=Prompt without context for instruction-tuned LMs.]
<|im_start|>system
You are Qwen, created by Alibaba Cloud. You are a helpful assistant.<|im_end|>
<|im_start|>user
Complete the following sentence. Only answer with the next word.
(*\bfseries <prompt>*)<|im_end|>
<|im_start|>assistant
\end{lstlisting}

\begin{lstlisting}[frame=single, title=Prompt with context for instruction-tuned LMs.]
<|im_start|>system
You are Qwen, created by Alibaba Cloud. You are a helpful assistant.<|im_end|>
<|im_start|>user
Complete the following sentence. Only answer with the next word.
Fact: (*\bfseries <context>*)
(*\bfseries <prompt>*)<|im_end|>
<|im_start|>assistant
\end{lstlisting}

\subsection{NQ}
We retain all samples from the development set of NQ\footnote{\url{https://console.cloud.google.com/storage/browser/natural_questions/v1.0/dev}} for which a short answer of fewer than five tokens is identified in the raw HTML of the corresponding Wikipedia pages.
Samples from the NQ dataset can be found in \Cref{tab:nq-contexts}.

\begin{table*}[h]
    \centering
    \scriptsize
    \begin{tabular}{p{2cm}lp{8.5cm}rl}
    \toprule
    Question & Short answer & Context & \#char. & Type \\
    \midrule
        when did the movie napoleon dynamite come out? & June 11, 2004 & <Table> <Tr> <Th colspan="2"> Napoleon Dynamite </Th> </Tr> <Tr> <Td colspan="2"> Theatrical release poster </Td> </Tr> <Tr> <Th> Directed by </Th> <Td> Jared Hess </Td> </Tr> <Tr> <Th> Produced by </Th> <Td> <Ul> <Li> Jeremy Coon </Li> <Li> Chris Wyatt </Li> <Li> Sean Covel </Li> <Li> Jory Weitz </Li> </Ul> </Td> </Tr> <Tr> <Th> Screenplay by </Th> <Td> <Ul> <Li> Jared Hess </Li> <Li> Jerusha Hess </Li> </Ul> </Td> </Tr> <Tr> <Th> Based on </Th> <Td> Peluca by Jared Hess </Td> </Tr> <Tr> <Th> Starring </Th> <Td> <Ul> <Li> Jon Heder </Li> <Li> Jon Gries </Li> <Li> Efren Ramirez </Li> <Li> Tina Majorino </Li> <Li> Aaron Ruell </Li> <Li> Diedrich Bader </Li> <Li> Haylie Duff </Li> </Ul> </Td> </Tr> <Tr> <Th> Music by </Th> <Td> John Swihart </Td> </Tr> <Tr> <Th> Cinematography </Th> <Td> Munn Powell </Td> </Tr> <Tr> <Th> Edited by </Th> <Td> Jeremy Coon </Td> </Tr> <Tr> <Th> Production company </Th> <Td> <Ul> <Li> MTV Films </Li> <Li> Napoleon Pictures </Li> <Li> Access Films </Li> </Ul> </Td> </Tr> <Tr> <Th> Distributed by </Th> <Td> <Ul> <Li> Fox Searchlight Pictures (North America) </Li> <Li> Paramount Pictures (International) </Li> </Ul> </Td> </Tr> <Tr> <Th> Release date </Th> <Td> <Ul> <Li> January 17, 2004 (2004 - 01 - 17) (Sundance) </Li> <Li> \textbf{June 11, 2004} (2004 - 06 - 11) (United States) </Li> <Li> </Li> <Li> </Li> <Li> </Li> </Ul> </Td> </Tr> <Tr> <Th> Running time </Th> <Td> 95 minutes </Td> </Tr> <Tr> <Th> Country </Th> <Td> United States </Td> </Tr> <Tr> <Th> Language </Th> <Td> English </Td> </Tr> <Tr> <Th> Budget </Th> <Td> \$400,000 </Td> </Tr> <Tr> <Th> Box office </Th> <Td> \$46.1 million </Td> </Tr> </Table> & 1,653 & Gold \\
        when was the lupus foundation of america founded? & 1977 & <P> The Lupus Foundation of America (LFA), founded in \textbf{1967}, is a national voluntary health organization based in Washington, D.C. with a network of chapters, offices and support groups located in communities throughout the United States . The Foundation is devoted to solving the mystery of lupus, one of the world's cruelest, most unpredictable and devastating diseases, while giving caring support to those who suffer from its brutal impact . Its mission is to improve the quality of life for all people affected by lupus through programs of research, education, support and advocacy . </P> & 592 & Conflicting \\
        who has scored the most tries in rugby union? & Daisuke Ohata & <P> This is a list of the leading try scorers in rugby union test matches . It includes players with a minimum of 30 test tries . </P> & 134 & Irrelevant \\
    \bottomrule
    \end{tabular}
    \caption{NQ samples and corresponding context types. `\#char.' indicates the length of the context, measured by number of characters.}
    \label{tab:nq-contexts}
\end{table*}

\paragraph{Sampling of conflicting contexts}
We create conflicting contexts that promote a different answer simply by taking the gold context and substituting the gold answer in the context. The substitute answer is sampled to yield coherent conflicting contexts, and to have a different meaning compared to the gold answer.

For a given question, context and short answer, we perform the following steps to identify substitute answers for conflicting contexts:
\begin{enumerate}[noitemsep]
    \item Check if the short answer is a date\footnote{Using the \texttt{dateutil.parser} in Python.}. If so, sample a new random date in the interval [1900, 2030) and format it in the same way as the gold date.
    \item If the short answer is not a date, prompt an LLM\footnote{The Cohere model \texttt{command-r-plus-08-2024} from \url{https://docs.cohere.com/v2/docs/command-r-plus}.} with the question and short answer to provide a substitute answer of the same format. If the proposed answer is already found in the sample context, prompt the model, for a maximum of 20 times, to generate another answer until a substitute answer not already found in the context has been generated.
\end{enumerate}
The prompt used to query an LLM for a substitute answer was as follows:
\begin{lstlisting}[frame=single, title=Prompt for getting substitute answers.]
## Instructions
Please provide an incorrect answer to the example below. 
The incorrect answer should be incorrect in the sense that it should be significantly different from the original answer. At the same time, it should be a plausible answer to the given question.
The incorrect answer should follow the same formatting as the original answer such that it should be possible to directly replace the original answer with the incorrect answer in any context.
The incorrect answer should be a single word or a short phrase.
Only output the incorrect answer.
    
## Example
Question: (*\bfseries <question>*)
Original answer:(*\bfseries <target\_true>*)
Incorrect answer:
\end{lstlisting}
In the event that the model generated a substitute answer that already could be found in the context, the previous model answer was added to the chat history together with the following new user query:

\begin{lstlisting}[frame=single, title=Prompt for getting another substitute answer.]
Please provide another incorrect answer following the same format as the original answer. Only output the incorrect answer.
\end{lstlisting}

\paragraph{Quality of conflicting contexts}
A manual inspection of 200 samples found the method reliable for producing adequate conflicting contexts with an accuracy of 90\% (11 samples corresponded to poor formatting, 4 were too similar to gold, and 4 were dropped due to data formatting issues or the LLM being unable to generate a substitute answer not already found in the context). In addition, we inspect the \textsc{CUB} results to ascertain the quality of the conflicting context sampling, see \Cref{app:results}.

We also experimented with a method based on named entities and random sampling for producing substitute answers for the conflicting contexts. In the method, the entity type of the answer to be replaced was detected and another named entity of the same type was randomly sampled from a NE dataset as the replacement. We found this method to work poorly compared to the LLM based approach. Mainly because the detected NEs lacked sufficient information for a successful sampling of replacements (e.g. ``2024'' and ``last year'' may both be labelled as time entities, while they are not interchangeable in all contexts).  

\paragraph{Sampling of irrelevant contexts}
Given a query and a corresponding Wikipedia page, the NQ annotators were instructed to mark the first paragraph in the Wikipedia page that contains an answer to the query. Therefore, to ensure that we only sample irrelevant contexts, we perform the sampling over all paragraphs before the gold paragraph in the given Wikipedia page.

We use the Jina Reranker v2\footnote{\texttt{jinaai/jina-reranker-v2-base-multilingual}} to identify the most relevant non-gold paragraph. It is a modern LM re-ranker that has been proven to work well on NQ \citep{hagstrom-etal-2025-language}.

\paragraph{Prompt templates}
The 2-shot prompts used to evaluate the LMs on NQ were as follows.
\begin{lstlisting}[frame=single, title=Prompt without context.]
Answer the following questions.
Question: When is the first episode of House of the Dragon released?
Answer: August 21, 2022

Question: In what country will the 2026 Winter Olympics be held?
Answer: Italy

Question: (*\bfseries <question>*)
Answer:
\end{lstlisting}
\begin{lstlisting}[frame=single, title=Prompt with context.]
Answer the following questions based on the context below.
Question: When is the first episode of House of the Dragon released?
Context: <Table> <Tr> <Th> Season </Th> <Th> Episodes </Th> <Th> First released </Th> <Th> Last released </Th> </Tr> <Tr> <Td> 1 </Td> <Td> 10 </Td> <Td> August 21, 2022 </Td> <Td> October 23, 2022 </Td> </Tr> <Tr> <Td> 2 </Td> <Td> 8 </Td> <Td> June 16, 2024 </Td> <Td> August 4, 2024
</Td> </Tr> </Table
Answer: August 21, 2022

Question: Where will the 2026 Winter Olympics be held?
Context: <P> The 2026 Winter Olympics (Italian: Olimpiadi invernali del 2026), officially the XXV Olympic Winter Games and commonly known as Milano Cortina 2026, is an upcoming international multi-sport event scheduled to take place from 6 to 22 February 2026 at sites across Lombardy and Northeast Italy. </P>
Answer: Lombardy and Northeast Italy

Question: (*\bfseries <question>*)
Context: (*\bfseries <context>*)
Answer:
\end{lstlisting}
For the instruction-tuned Qwen models, a chat template with slightly different prompt templates was used. The 2-shot prompt templates for the instruction-tuned models were as follows.
\begin{lstlisting}[frame=single, title=Prompt without context for instruction-tuned LMs.]
<|im_start|>system
You are Qwen, created by Alibaba Cloud. You are a helpful assistant.<|im_end|>
<|im_start|>user
Answer the question. Only answer with the answer. Examples of questions and desired answers are given below.

# Example 1
Question: When is the first episode of House of the Dragon released?
Answer: August 21, 2022

# Example 2
Question: In what country will the 2026 Winter Olympics be held?
Answer: Italy

# Now, answer the following question (only with the answer):
Question:(*\bfseries <question>*)
Answer:<|im_end|>
<|im_start|>assistant
\end{lstlisting}
\begin{lstlisting}[frame=single, title=Prompt with context for instruction-tuned LMs.]
<|im_start|>system
You are Qwen, created by Alibaba Cloud. You are a helpful assistant.<|im_end|>
<|im_start|>user
Answer the question based on the provided context. Only answer with the answer. Examples of questions and desired answers are given below.

# Example 1
Question: When is the first episode of House of the Dragon released?
Context: <Table> <Tr> <Th> Season </Th> <Th> Episodes </Th> <Th> First released </Th> <Th> Last released </Th> </Tr> <Tr> <Td> 1 </Td> <Td> 10 </Td> <Td> August 21, 2022 </Td> <Td> October 23, 2022 </Td> </Tr> <Tr> <Td> 2 </Td> <Td> 8 </Td> <Td> June 16, 2024 </Td> <Td> August 4, 2024
</Td> </Tr> </Table
Answer: August 21, 2022

# Example 2
Question: Where will the 2026 Winter Olympics be held?
Context: <P> The 2026 Winter Olympics (Italian: Olimpiadi invernali del 2026), officially the XXV Olympic Winter Games and commonly known as Milano Cortina 2026, is an upcoming international multi-sport event scheduled to take place from 6 to 22 February 2026 at sites across Lombardy and Northeast Italy. </P>
Answer: Lombardy and Northeast Italy

# Now, answer the following question (only with the answer):
Question: (*\bfseries <question>*)
Context: (*\bfseries <context>*)
Answer:<|im_end|>
<|im_start|>assistant
\end{lstlisting}

\subsection{DRUID}
We map the stances of DRUID to context type using the following approach:
\begin{enumerate}[noitemsep,topsep=0pt]
    \item Gold: If the evidence is relevant and the stance of the evidence aligns with the claim verdict reached by the fact-check site (here considered gold). This automatically encompasses most samples with evidence that has been sampled from a fact-check site, as the stance of the evidence is likely to align with the FC verdict.
    \item Conflicting: If the evidence is relevant and the stance of the evidence does not align with the claim verdict. This automatically encompasses all samples with insufficient evidence, as the original FC verdicts always are True, Half True or False.
    \item Irrelevant: If the evidence is irrelevant.
\end{enumerate}
Samples from the DRUID dataset can be found in \Cref{tab:druid-contexts}. The evidence stance and fact-check verdict distributions per context type can be found in \Cref{tab:druid-stance,tab:druid-fc-verdict}.

\begin{table*}[ht]
    \centering
    \scriptsize
    \resizebox{\textwidth}{!}{
    \begin{tabular}{p{1cm}p{2cm}lp{9.3cm}rl}
    \toprule
    Claimant & Claim & Verdict & Evidence & \#char. & Type \\
    \midrule
     Viral Claim & Harvard professor Charles Lieber was arrested for manufacturing and selling the new coronavirus to China & False & Lieber was arrested on January 28 for "making false statements to the agency of the United States Government," or lying to federal authorities about his ties to China, as per the fact-check report. The channel added that prosecutors have never alleged that Lieber was involved in manufacturing and/or selling a virus to China. The full federal court complaint against Dr Lieber can be read <a href="https://htv-prod-media.s3.amazonaws.com/files/lieber-complaint-1586387800.pdf" rel="noopener noreferrer" target="\_blank">here</a>.</p>.<p>The report also clarified Lieber’s links to Wuhan. The report stated, "Lieber travelled to WUT (Wuhan University of Technology) in mid-November 2011 ostensibly in order to participate in a Nano-Energy Materials Forum."</p>.<p>On July 29, Dr Lieber’s attorney Marc Mukasey told WCVB Channel 5 that he didn’t hide anything or get paid as the government alleges.</p>.<p>Thus, the social media claim that Harvard professor Dr Charles Lieber "made and sold" the Covid-19 virus to China is false.</p> & 1,032 & Gold \\
     FACEBOOK POST & WikiLeaks has published the 1st list of black money holders in Swiss banks. & False & (See attached file: List of Black Money Holders from Wiki & 57 & Conflict. \\
     Irish Congress of Trade Unions (ICTU) & One in five school staff in Northern Ireland are assaulted at least once a week. & False & Finnegan, who died in January 2002, had also abused boys at St. Colman’s College, a prestigious Catholic boys’ secondary school in Newry, Northern Ireland. He taught there from 1967 to 1971 and again from 1973 to 1976, when he was appointed president of the school. He served in that post until 1987. [...] Admitted on October 9, 2014 to sample charges of indecently assaulting four boys as young as 10 at St Mary’s CBS primary school in Mullingar between 1984 and 1987. Jailed for two years at Mullingar Circuit Court sitting in Tullamore. This concluded a ten-year investigation by detectives in Mullingar. [...] When Smyth returned to Kilnacrott in 1983, he again began abusing children in Belfast, including the girl who, on February 23, 1990, would meet with a social worker at the Catholic Family Welfare Society in Belfast and start all the Smyth revelations. & 866 & Irrel. \\
    \bottomrule
    \end{tabular}
    }
    \caption{DRUID samples and corresponding context types. `Conflict.' and `Irrel.' denote conflicting and irrelevant context types, respectively. `\#char.' indicates the length of the evidence (context), measured by number of characters.}
    \label{tab:druid-contexts}
\end{table*}

\begin{table}[h]
    \centering
    \scriptsize
    \begin{tabular}{l l r}
    \toprule
    Context     & Evidence stance & Count \\
    \midrule
        Gold & Refutes & 1,579 \\
        & Supports & 359 \\
        Conflicting & Refutes & 35 \\
        & Insufficient-refutes & 437 \\
        & Insufficient-contradictory & 163 \\
        & Insufficient-neutral & 892 \\
        & Insufficient-supports & 585 \\
        & Supports & 367 \\
        Irrelevant & \texttt{not applicable} & 83 \\
    \bottomrule
    \end{tabular}
    \caption{Stance distribution per context type for DRUID.}
    \label{tab:druid-stance}
\end{table}

\begin{table}[h]
    \centering
    \scriptsize
    \begin{tabular}{l l r}
    \toprule
    Context     & FC verdict & Count \\
    \midrule
        Gold & False & 1,579 \\
        & True & 359 \\
        Conflicting & False & 1,842 \\
        & Half True & 276 \\
        & True & 361 \\
        Irrelevant & False & 54 \\
        & Half True & 13 \\
        & True & 16 \\
    \bottomrule
    \end{tabular}
    \caption{Fact-check verdict distribution per context type for DRUID.}
    \label{tab:druid-fc-verdict}
\end{table}

Differently from CounterFact and NQ, no context synthesis is necessary for the DRUID samples as they, by virtue of utilising naturally occurring samples from a RAG pipeline, already contain samples representative of gold, conflicting and irrelevant contexts. 

\paragraph{Prompt templates}
The 2-shot prompts used for evaluating the LMs on DRUID were as follows.
\begin{lstlisting}[frame=single, title=Prompt without context.]
Are the following claims True or False? Answer None if you are not sure or cannot answer.

Claimant: Viral post
Claim: "the new coronavirus has HIV proteins that indicate it was genetically modified in a laboratory."
Answer: False

Claimant: Sara Daniels
Claim: "Blackpink released the single 'You me too' in 2026."
Answer: None

Claimant: (*\bfseries <claimant>*)
Claim: "(*\bfseries <claim>*)"
Answer:
\end{lstlisting}
\begin{lstlisting}[frame=single, title=Prompt with context.]
Are the claims True or False based on the accompanying evidence? If you are not sure or cannot answer, say None.

Claimant: Viral post
Claim: "the new coronavirus has HIV proteins that indicate it was genetically modified in a laboratory."
Evidence: "Microbiologists say the spike proteins found in the new coronavirus are different from the ones found in HIV. [...] There is no evidence to suggest the coronavirus was genetically modified."
Answer: False

Claimant: Sara Daniels
Claim: "Blackpink released the single 'You me too' in 2026."
Evidence: "Blackpink released their album 'Born Pink' in 2022."
Answer: None

Claimant: (*\bfseries <claimant>*)
Claim: "(*\bfseries <claim>*)"
Evidence: "(*\bfseries <evidence>*)"
Answer:
\end{lstlisting}
For the instruction-tuned Qwen models, a chat template with slightly different prompt templates was used for compatibility. The 2-shot prompt templates for the instruction-tuned models were as follows.
\begin{lstlisting}[frame=single, title=Prompt without context for instruction-tuned LMs.]
<|im_start|>system
You are Qwen, created by Alibaba Cloud. You are a helpful assistant.<|im_end|>
<|im_start|>user
Is the claim True or False? Answer None if you are not sure or cannot answer. Only answer with True, False or None. Examples of claims and desired answers are given below.

# Example 1
Claimant: Viral post
Claim: "the new coronavirus has HIV proteins that indicate it was genetically modified in a laboratory."
Answer: False

# Example 2
Claimant: Sara Daniels
Claim: "Blackpink released the single 'You me too' in 2026."
Answer: None

# Now, answer for the following claim:
Claimant: (*\bfseries <claimant>*)
Claim: "(*\bfseries <claim>*)"
Answer (True, False or None):<|im_end|>
<|im_start|>assistant
\end{lstlisting}
\begin{lstlisting}[frame=single, title=Prompt with context for instruction-tuned LMs.]
<|im_start|>system
You are Qwen, created by Alibaba Cloud. You are a helpful assistant.<|im_end|>
<|im_start|>user
Is the claim True or False based on the accompanying evidence? If you are not sure or cannot answer, say None. Only answer with True, False or None. Examples of claims, evidence and desired answers are given below.

# Example 1
Claimant: Viral post
Claim: "the new coronavirus has HIV proteins that indicate it was genetically modified in a laboratory."
Evidence: "Microbiologists say the spike proteins found in the new coronavirus are different from the ones found in HIV. [...] There is no evidence to suggest the coronavirus was genetically modified."
Answer: False

# Example 2
Claimant: Sara Daniels
Claim: "Blackpink released the single 'You me too' in 2026."
Evidence: "Blackpink released their album 'Born Pink' in 2022."
Answer: None

# Now, answer for the following claim:
Claimant: (*\bfseries <claimant>*)
Claim: "(*\bfseries <claim>*)"
Evidence: "(*\bfseries <evidence>*)"
Answer (True, False or None):<|im_end|>
<|im_start|>assistant
\end{lstlisting}

\subsection{Context Lengths}
\label{app:context_lengths}
\begin{figure}[h]
\centering
\begin{subfigure}[t]{0.8\linewidth}
\includegraphics[width=\linewidth]{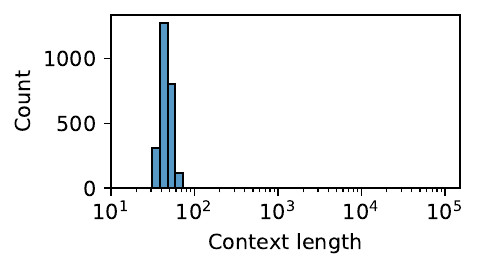} 
\caption{CounterFact}
\label{fig:cl-counterfact}
\end{subfigure}
\begin{subfigure}[t]{0.8\linewidth}
\includegraphics[width=\linewidth]{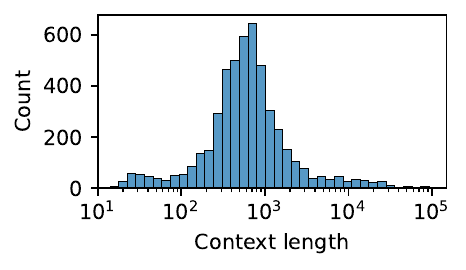}
\caption{NQ}
\label{fig:cl-nq}
\end{subfigure}
\begin{subfigure}[t]{0.8\linewidth}
\includegraphics[width=\linewidth]{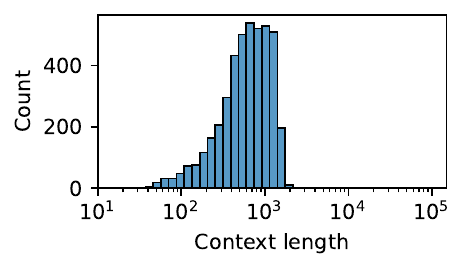}
\caption{DRUID}
\label{fig:cl-druid}
\end{subfigure}
\caption{Context lengths of the test splits of the datasets used in \textsc{CUB}, measured as number of characters. The mean context lengths are 46, 1810, 700 for CounterFact, NQ and DRUID, respectively.}
\label{fig:context-length}
\end{figure}

We measure context lengths (number of characters) for each dataset in \textsc{CUB} and report the results in \Cref{fig:context-length}. Data examples and corresponding context lengths can be found in \Cref{tab:counterfact-examples,tab:nq-contexts,tab:druid-contexts}, for comparison.

\section{CCU metric}\label{app:metrics}

$\mathrm{BCU}$ cannot measure the difference in model behaviour when context is introduced, as it does not take model behaviour without context into consideration. To address this, we introduce $\mathrm{CCU}$. Given a query $Q$ and context $C$, $\mathrm{CCU}$ measures the change in probability for token $t$ as follows.
\begin{equation}
    \mathrm{CCU}(t) = 
\begin{cases}
    \frac{P_M(t|Q,C)-P_M(t|Q)}{1-P_M(t|Q)} \\ 
    \text{\small if } \scriptstyle P_M(t|Q,C)\geq P_M(t|Q), \vspace{0.3cm}\\
    \frac{P_M(t|Q,C)-P_M(t|Q)}{P_M(t|Q)} \\ 
    \text{\small otherwise.}
\end{cases}
\label{eq:ccu}
\end{equation}
For relevant contexts $C$ we record $\mathrm{CCU}(t_C)$, i.e. the scores for the token promoted by the context. For irrelevant contexts we record the $\mathrm{CCU}(t_M)$, i.e. the scores for the top token predicted by the model when prompted without context (memory). The range of $\mathrm{CCU}$ is $[-1, 1]$, for which a value of $-1$ denotes that the model goes completely \emph{against} the context when the context is relevant or against its memory when the context is irrelevant, and vice versa for $\mathrm{CCU}$ values of 1. We report the averaged $\mathrm{CCU}$ per context type.

By measuring the token probabilities before and after context is introduced, the $\mathrm{CCU}$ metric more accurately captures how the LM is impacted by context. However, this metric excludes the API-based models, which do not provide the output logits necessary to compute $\mathrm{CCU}$ scores.

\section{Hyperparameter Search}
\label{app:hyperparameter_search}

\subsection{Prompting}
The name of the tuned prompt found for each model and dataset can be found in \Cref{tab:tuned-prompts}. Different sets of prompts were experimented with depending on dataset and model type. A set of 11 to 12 prompts were produced for each of CounterFact, NQ and DRUID for the three different model types (causal LM, instruction-tuned LMs and API-based models), respectively. Prompts with the same number are similar to each other across model types (e.g. Prompt \#2 for Qwen2.5 on DRUID is similar to Prompt \#2 for instruction-tuned Qwen2.5 on DRUID). Prompt sets across different datasets are dissimilar as they are adapted to align the instructions and few-shot examples to the given dataset. Prompt sets across different model types for the same dataset are dissimilar as small tweaks need to be applied for the instruction-tuned models that work less well in a purely causal language modelling setting, and for API-based models that are chat-oriented. All prompts are possible to view in the code repository of the paper.

\input{Tables/tuned_prompts.tex}

\subsection{PH3}
The tuned attention head configurations for PH3 can be found in \Cref{tab:ph3-heads}. The head configurations are grouped by the top number of identified attention heads to consider and to what extent we allow mixing between context and memory heads. E.g. \#25 \texttt{all} denotes all top-25 context and memory heads detected, \#3 \texttt{memory} denotes the top-3 memory heads, allowing for overlap with context heads, and \#1 \texttt{only memory} denotes memory heads detected without overlap with context heads when considering the top-1 context and memory heads.

\begin{table*}[h!]
    \centering
    \scriptsize
    \begin{tabular}{ll|p{3.7cm}p{3.7cm}p{3.7cm}}
    \toprule
    Model & Mode & CounterFact & NQ & DRUID \\
    \midrule
    {\scshape GPT2-XL} & +context & \#25 \texttt{all} & \#1 \texttt{all} & \#5 \texttt{only memory}* \\
    & & L18H10, L21H10, L21H7, L22H18, L22H20, L24H6, L26H14, L26H20, L26H8, L27H15, L27H5, L28H15, L29H5, L29H9, L30H21, L30H8, L31H0, L31H3, L31H8, L32H13, L33H14, L33H18, L33H2, L33H7, L34H17, L34H20, L35H17, L35H19, L35H21, L36H17, L36H2, L37H7, L38H24, L38H7, L39H12, L39H9, L40H13, L40H23, L41H5, L41H9, L42H24, L43H15, L47H0 & L28H15, L35H19 & L32H13, L35H19, L42H24, L43H15 \\
    & +memory & \#12 \texttt{memory} & \#7 \texttt{only context} & \#22 \texttt{all} \\
    & & L26H14, L26H8, L32H13, L33H14, L35H19, L38H24, L40H23, L41H5, L42H24, L43H15, L47H0, L30H8 & L27H15, L28H15, L29H9, L33H2, L34H17, L37H7 & L21H10, L22H20, L24H6, L26H14, L26H20, L26H8, L27H15, L27H5, L28H15, L29H9, L30H21, L30H8, L31H0, L31H3, L31H8, L32H13, L33H14, L33H18, L33H2, L33H7, L34H17, L34H20, L35H17, L35H19, L36H17, L36H2, L37H7, L38H24, L38H7, L39H12, L39H9, L40H13, L40H23, L41H5, L42H24, L43H15, L47H0 \\
    \midrule
    {\scshape Pythia 6.9B} & +context & \#15 \texttt{memory} & \#17 \texttt{only memory} & \#10 \texttt{only context} \\
    & & L10H27, L14H6, L16H16, L17H28, L19H11, L19H21, L20H11, L20H18, L21H8, L27H22, L18H7, L19H28, L20H2, L20H8, L24H5 & L10H27, L14H28, L14H6, L16H16, L17H28, L19H11, L19H21, L20H11, L20H18, L21H8, L22H12, L27H22 & L12H11, L12H13, L14H0, L15H17, L17H14, L20H2, L8H11 \\
    & +memory & \#25 \texttt{only context} & \#12 \texttt{only context} & \#17 \texttt{only context} \\
    & & L10H1, L12H11, L12H13, L13H12, L14H0, L14H23, L15H17, L17H14, L18H10, L19H1, L19H20, L21H10, L23H25, L29H22, L8H11, L8H24 & L12H11, L12H13, L14H0, L14H23, L15H17, L17H14, L19H31, L20H2, L8H11 & L10H1, L12H11, L12H13, L13H12, L14H0, L14H23, L15H17, L17H14, L18H10, L19H1, L19H31, L8H11 \\
    \midrule
    {\scshape Qwen2.5 1.5B} & +context & \#15 \texttt{only memory} & \#12 \texttt{only memory} & \#17 \texttt{only context} \\
    & & L10H0, L10H1, L13H1, L16H1, L17H0, L18H0, L1H1, L3H0 & L10H0, L13H1, L16H1, L17H0, L18H0, L1H1 & L14H1, L16H0, L18H1, L19H0, L19H1, L20H1, L24H1, L26H0, L26H1, L9H0 \\
    & +memory & \#5 \texttt{only context} & \#12 \texttt{only context} & \#12 \texttt{only memory} \\
    & & L15H1, L16H0, L27H0 & L14H1, L16H0, L18H1, L19H0, L24H1, L27H0 & L10H0, L13H1, L16H1, L17H0, L18H0, L1H1 \\
    \midrule
    {\scshape Qwen2.5 1.5B} & +context & \#7 \texttt{only memory} & \#1 \texttt{only context} & \#10 \texttt{only context} \\
    \emph{Instruct} & & L15H0, L1H1, L21H0 & L19H1 & L14H0, L17H1, L19H1, L22H0, L26H0 \\
    & +memory & \#1 \texttt{only context} & \#12 \texttt{only context}* & \#5 \texttt{only context} \\
    & & L19H1 & L14H0, L17H1, L19H1, L22H0, L26H0, L27H0 & L17H0, L19H1, L22H0 \\
    \midrule
    {\scshape Qwen2.5 7B} & +context & \#7 \texttt{memory} & \#1 \texttt{only context} & \#3 \texttt{only memory} \\
    & & L0H0, L17H1, L18H2, L19H0, L21H0, L22H2, L23H0 & L27H0 & L0H0, L22H2 \\
    & +memory & \#15 \texttt{only context} & \#5 \texttt{only context} & \#12 \texttt{only context} \\
    & & L13H0, L17H0, L18H1, L18H3, L22H0, L24H3, L25H1, L26H0, L27H0, L27H2 & L22H0, L27H0, L27H2 & L16H3, L17H0, L18H1, L18H3, L22H0, L24H3, L26H0, L27H0, L27H2 \\
    \midrule
    {\scshape Qwen2.5 7B} & +context & \#17 \texttt{only memory} & \#5 \texttt{context} & \#5 \texttt{only context} \\
    \emph{Instruct} & & L11H1, L12H0, L13H3, L14H3, L16H1, L17H0, L17H3, L18H2, L1H1, L20H0, L21H2, L26H3, L3H0 & L18H0, L18H3, L22H2, L23H0, L27H2 & L18H0, L18H3, L27H2 \\
    & +memory & \#3 \texttt{only context} & \#3 \texttt{only context} & \#17 \texttt{all} \\
    & & L18H0 & L18H0 & L0H0, L11H1, L12H0, L13H3, L14H3, L15H1, L16H0, L16H1, L17H0, L17H3, L18H0, L18H1, L18H2, L18H3, L19H0, L19H3, L1H1, L20H0, L20H2, L20H3, L21H0, L21H2, L22H0, L22H2, L23H0, L26H3, L27H0, L27H2, L3H0, L8H1 \\
    \bottomrule
    \end{tabular}
    \caption{Tuned PH3 attention head configurations for each model and evaluation dataset. +context indicates heads for which pruning leads to increased context usage and vice versa for +memory. Configurations marked with * denote that they yielded degraded performance compared to the standard setting (no mechanistic intervention) on the validation set.}
    \label{tab:ph3-heads}
\end{table*}

\subsection{Context-aware Contrastive Decoding: COIECD}

\begin{table}[ht]
    \centering
    \scriptsize
    \begin{tabular}{l|cc}
    \toprule
    Model & $\lambda$ & $\alpha$ \\
    \midrule
    {\scshape GPT2-XL} & 0.50 & 1.00 \\
    {\scshape Pythia 6.9B} & 0.50 & 1.00 \\
    {\scshape Qwen2.5 1.5B} & 1.00 & 0.50 \\
    {\scshape Qwen2.5 1.5B Instruct} & 0.50 & 1.00 \\
    {\scshape Qwen2.5 7B} & 1.00 & 1.00 \\
    {\scshape Qwen2.5 7B Instruct} & 0.25 & 0.50 \\
    {\scshape Qwen2.5 32B} & 0.50 & 1.00 \\
    {\scshape Qwen2.5 32B Instruct} & 0.50 & 1.50 \\
    \bottomrule
    \end{tabular}
    \caption{Selected COIECD hyperparameters $\lambda$ and $\alpha$ for each model, evaluated on gold contexts from NQ's validation set. For models with multiple $(\lambda, \alpha)$ pairs attaining the maximum score, we choose the setting that lies near the midpoint of the optimal region.}
    \label{tab:coiecd_hp}
\end{table}

Unlike other CMTs, the hyperparameters used in COIECD, $\alpha$ and $\lambda$, are selected following the original paper, \citet{yuan-etal-2024-discerning}, using the gold context from the validation set of NQ dataset.
This deviation is necessary, as optimising COIECD's hyperparameters by maximising the average BCU across all context types causes the model to converge to using only the output distribution without context in the decoding step.
This outcome arises from the nature of COIECD, where always relying on the distribution without context results in a BCU score of 1.0 for irrelevant contexts, while also causing the model to ignore context, including gold and conflicting contexts.
To prevent COIECD from collapsing into regular generation without context and to enable meaningful comparison with other CMTs, we follow the hyperparameter search from the original paper.
While \citet{yuan-etal-2024-discerning} uses the same hyperparameter values across all models, our models exhibit different tendencies during hyperparameter search.
Therefore, we tune the hyperparameters separately for each model to ensure a fair comparison with other methods.
We search $\alpha$ in the range [0.0, 2.0] and $\lambda$ in the range [0.1, 1.0], and the hyperparameters for each model are in Table \ref{tab:coiecd_hp}.

\begin{table}[h]
\resizebox{\columnwidth}{!}{%
\begin{tabular}{@{}lrlr@{}}
\toprule
\textbf{Dataset}           & \textbf{Dataset weight} & \multicolumn{1}{c}{\textbf{Context type}} & \textbf{Context weight} \\ \midrule
\multirow{3}{*}{SQuAD 2.0} & \multirow{3}{*}{0.4}    & Relevant                                  & 0.65                    \\
                           &                         & Irrelevant                                & 0.25                    \\
                           &                         & Empty                                     & 0.1                     \\ \midrule
\multirow{3}{*}{TriviaQA}  & \multirow{3}{*}{0.3}    & Relevant                                  & 0.65                    \\
                           &                         & Irrelevant                                & 0.25                    \\
                           &                         & Empty                                     & 0.10                     \\ \midrule
\multirow{3}{*}{AVeriTeC}  & \multirow{3}{*}{0.15}   & Relevant                                  & 0.65                    \\
                           &                         & Irrelevant                                & 0.25                    \\
                           &                         & Empty                                     & 0.10                     \\ \midrule
\multirow{4}{*}{DYNAMICQA} & \multirow{4}{*}{0.15}   & Relevant                                  & 0.50                     \\
                           &                         & Irrelevant                                & 0.05                    \\
                           &                         & Empty                                     & 0.05                    \\
                           &                         & Counterfactual                            & 0.40                     \\ \bottomrule
\end{tabular}%
}
\caption{Sampling weight for each dataset. We first sample the number of instances for each dataset following the dataset sampling weight. Then, each context type is determined by the context sampling weight.}
    \label{tab:ft_dataset}
\end{table}

\section{Implementation Details of Fine-tuning}\label{app:finetuning}

To align the domain with our evaluation data, we curate the fine-tuning data with two QA datasets \cite{joshi-etal-2017-triviaqa, rajpurkar-etal-2018-know}, one FC dataset \cite{averitec}, and one sentence completion dataset \cite{marjanovic-etal-2024-dynamicqa}. Before fine-tuning each LM, we elicit its parametric answers by querying without contexts. We then select the questions that the LM answered correctly and pair them with irrelevant and empty contexts. The fine-tuning data thus contains contexts that can be irrelevant, counterfactual, or empty. During fine-tuning, we train the LM to generate answers aligned with the provided context. When the context is irrelevant, we train the LM to be robust, i.e. ignore the context and output its parametric answer. Due to the computational costs associated with fine-tuning billion-sized LMs, we use the Low-Rank Adaptation method \cite{lora}.

The LMs are fine-tuned with a learning rate of 5e-5,\footnote{Experiments with other learning rates yielded insignificant changes in performance on the validation set.} using warm-up. To avoid overfitting, we use early stopping based on the loss on the validation set. For QA datasets, we use the train split from SQuAD 2.0 \cite{rajpurkar-etal-2018-know}, and TriviaQA \cite{joshi-etal-2017-triviaqa}. For a FC dataset, we take the train split from AVeriTeC \cite{averitec}. For a sentence completion dataset, we take the static partition of the DYNAMICQA \cite{marjanovic-etal-2024-dynamicqa}. We only create counterfactual training examples with DYNAMICQA dataset. The detailed statistics for mixing the selected datasets can be found in \Cref{tab:ft_dataset}.

\section{Additional Details of Multi-agent} \label{appendix:multiagent}

\begin{figure}
    \centering
    \includegraphics[width=1\columnwidth]{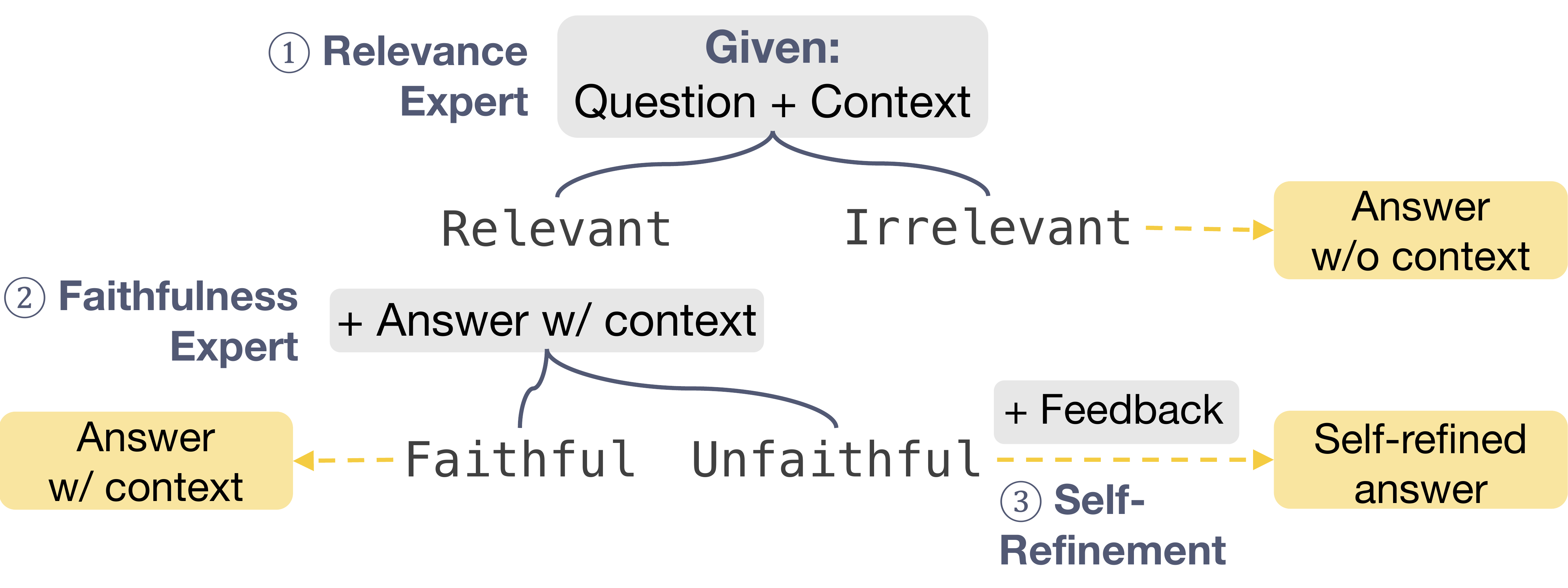}
    \caption{Overview of the multi-agent approach.}
    \label{fig:multi_agent}
\end{figure}

\begin{algorithm}
\caption{Multi-agent}
\label{alg:multi_agent}
    \begin{algorithmic}[1]
    \State \textbf{Given:} question $q$, context $c$
    \State \textbf{Stage1: Relevance Assessment}
    \State Predict $f_{\text{rel}} \sim \text{LM}_{\text{rel}}(f_{\text{rel}} \mid q, c)$
    \If{$f_{\text{rel}} = \texttt{Relevant}$}
        \State Proceed to Stage 2
    \Else
        \State \Return $\text{LM}(a \mid q)$ \Comment{Answer w/o $c$}
    \EndIf
    \State \textbf{Stage 2: Context-Faithfulness}
    \State Predict $a_{c} \sim \text{LM}(a_{c} \mid q,c)$
    \State Predict $f_{\text{faith}} \sim \text{LM}_{\text{faith}}(f_{\text{faith}} \mid q, c, a_{c})$
    
    \If{$f_{\text{faith}} = \texttt{Faithful}$}
        \State \Return $a_c$ \Comment{Answer w/ $c$} 
    \Else
        \State Proceed to Stage 3
    \EndIf
    \State \textbf{Stage 3: Self-Refinement}
    \State \Return $\text{LM}(a \mid q, c, a_{c}, f_{\text{faith}})$ \Comment{Self-Refined}
    \end{algorithmic}
\end{algorithm}

As illustrated in the algorithm and Figure \ref{fig:multi_agent}, we first assess relevance using the relevance agent to determine whether the provided context should be used.
Then, the faithfulness agent provides feedback on the model response that was generated with context.
If the feedback indicates that the initial answer is unfaithful, the model generates a self-refined answer based on that feedback.
Given that these tasks require instruction-following capabilities, we restrict our evaluation to instruction-tuned or chat LMs.

We design the \texttt{Multi-agent} approach to investigate whether LMs can explicitly handle the two objectives of context utilisation: (1) being robust to irrelevant context and (2) being faithful to relevant context.
Rather than directly generating an answer, an LM is guided to perform intermediate reasoning steps, each handled by a dedicated LM agent.
This decomposition allows us to understand whether LMs can explicitly recognise when the context should be used and whether their answer aligns with it when it is.
While self-refinement and LM agent have been used broadly in reasoning tasks \citep{10.5555/3692070.3692537, feng-etal-2024-dont, madaan2023selfrefine}, our motivation is grounded in examining two components of context utilisation separately.
Notably, self-refinement is only applied when the context is assessed as relevant but the answer is assessed as unfaithful, reflecting our focus on improving the usage of relevant context.
By structuring the problem in this way, we aim to better understand the extent to which LMs can reason about context relevance and faithfulness.

Figure \ref{fig:multi_agent} and Algorithm \ref{alg:multi_agent} outline the \texttt{Multi-agent} procedure employed in our framework.
Given a question and the context, the model first undergoes a relevance assessment stage, where it is explicitly instructed to determine whether the context is relevant to the question \citep{shen-etal-2024-assessing}. 
If assessed as irrelevant, the model answers without the context; if relevant, it incorporates the context to generate the initial answer and proceeds to the next stage.
In the context faithfulness assessment, the model is instructed to provide feedback on whether its answer faithfully reflects the provided context.
If deemed faithful, the answer is retained as the final answer.
If the prediction is assessed as unfaithful, the model is instructed to refine its answer using the question, context, initial answer, and feedback derived from the faithfulness assessment.
This self-refinement stage encourages the model to self-correct based on its own feedback.
To ensure consistency in output formatting during refinement, we incorporate two-shot demonstrations.

The templates for relevance assessment, context faithfulness, and self-refinement are presented below. 
Task-specific templates for each dataset are available in the released code.

\begin{lstlisting}[frame=single, title=Relevance Assessment (NQ)]
You are a relevance assessment expert. Your task is to evaluate whether the provided context is relevant to the question.

Context: {context}
Question: {question}

If the provided context is relevant to the question, answer "Relevant", otherwise answer "Irrelevant". Do not rely on your own knowledge or judge the factual accuracy of the context.
Answer:
\end{lstlisting}

\begin{lstlisting}[frame=single, title=Context faithfulness (CounterFact and NQ)]
You are a context-faithfulness expert. Your task is to evaluate whether the proposed answer faithfully uses the information in the provided context.

Context: {context}
Question: {question}
Proposed answer: {response}

Does the answer faithfully reflect the content of the context? Do not rely on your own knowledge or judge the factual accuracy of the context. Please explain briefly.

Feedback:
\end{lstlisting}

\begin{lstlisting}[frame=single, title=Self-refinement (NQ)]
Your task is to generate the best possible final answer to the question, based on the expert feedback. 
You may keep the original proposed answer if it is correct, or revise it if the feedback suggests it is incorrect or unsupported.
Generate only the final answer. Do not include any explanation or repeat the prompt.

{Two demonstrations}

Context: {context}
Question: {question}
Proposed answer: {response}
Feedback on context faithfulness: {feedback}
Final answer:
\end{lstlisting}

\section{Additional Details on PH3}\label{app:ph3}

The PH3 method is implemented in two steps: 1) identification of attention heads responsible for context or memory reliance via path patching and 2) pruning the identified attention heads for increased memory or context usage. To identify attention heads, we use the CounterFact datasets with samples that elicit exact fact recall in each studied model \citep{saynova2025factrecallheuristicspure}. For the evaluation on our studied datasets, we tune the number of heads to prune on the validation splits of each evaluation dataset, similarly to the approach by \citet{jin-etal-2024-cutting}. The attention head configuration is tuned for each mode (\texttt{PH3 +memory} and \texttt{PH3 +context}, respectively).

\section{Input Features} \label{app:features}

We detect the input features described in \Cref{sec:input-features} as follows:
\begin{itemize}[noitemsep,topsep=0pt]
    \item Context length is measured by the number of characters in the context. 
    \item Flesch reading ease score is measured with the \texttt{textstat}\footnote{\url{https://github.com/textstat/textstat}} module.
    \item Query-context overlap is measured as the size of the set of words that form the intersection of the set of words in the query and context, respectively, normalised by the size of the set of query words. CounterFact is excluded from this analysis as its synthetic samples yield trivial results for this feature. 
    \item The answer position is measured as the index of the answer in the context normalised by context length. This feature is only detectable for gold and conflicting contexts for CounterFact and NQ.
    \item The distractor rate is measured as the number of answer entities found in the context, divided by the total number of entities in the context with an entity type that matches the answer entity type(s).\footnote{Named entities are detected using spaCy and \texttt{en\_core\_web\_trf}.} This feature is similarly only measurable for gold and conflicting contexts from CounterFact and NQ.
    \item Relevance is given by the relevance agent based on Qwen 32B Instruct from the Multi-agent setup. It labels context as either `relevant' or `irrelevant'.
\end{itemize}

\section{Computational Resources}
GPT2-XL was evaluated using one Nvidia T4 GPU. Pythia, Qwen 1.5B and Qwen 7B using one A40 GPU. Qwen 32B was evaluated using four A40 GPUs. The compute budget for all CMTs was about 14 hours per model for CounterFact, 28 hours per model for NQ and 21 hours per model for DRUID, amounting to a total of about 900 GPU hours.   

The costs for the experiments with Cohere Command A amounted to a total of about 120 USD. The costs for the experiments with OpenAI's GPT-4.1 and GPT-4.1 mini amounted to a total of about 110 USD.

\section{Use of AI assistants}

AI assistants like Copilot and ChatGPT were intermittently used to generate template code and rephrase sentences in the paper, etc. However, no complete paper sections or code scripts have been generated by an AI assistant. All generated content has been inspected and verified by the authors.

\end{document}

%% file: Tables/cmtsSummary.tex
\definecolor{Paired1}{RGB}{166,206,227}
\definecolor{Paired2}{RGB}{31,120,180}
\definecolor{Paired3}{RGB}{178,223,138}
\definecolor{Paired4}{RGB}{51,160,44}
\definecolor{Paired5}{RGB}{251,154,153}
\definecolor{Paired6}{RGB}{227,26,28}
\definecolor{Paired7}{RGB}{253,191,111}
\definecolor{Paired8}{RGB}{255,127,0}
\definecolor{Paired9}{RGB}{202,178,214}
\definecolor{Paired10}{RGB}{106,61,154}
\definecolor{Paired11}{RGB}{255,255,153}
\definecolor{Paired12}{RGB}{177,89,40}

\definecolor{Set2Gray}{RGB}{153,153,153}

\begin{table}[t]
\centering
\resizebox{\columnwidth}{!}{
\begin{tabular}{r|cccc}
\toprule
\textbf{Methods} & \textbf{Objective} & \textbf{Level} & \textbf{\shortstack[c]{Tuning \\ Cost}} & \textbf{\shortstack[c]{Inference \\ Cost}} \\
\midrule
\cellcolor{Paired5} \textcolor{black}{Fine-tuning}     & Both    & Fine-tuning  & High & Low  \\
\cellcolor{Paired7} \textcolor{black}{Prompting}   & Both    & Prompt.    & Low  & Low  \\
\cellcolor{Paired8} \textcolor{white}{Multi-agent}     & Both    & Prompt.    & None & High \\
\cellcolor{Paired1} \textcolor{black}{PH3 +context}   & Faith   & Mech. & High & Low  \\
\cellcolor{Paired2} \textcolor{white}{COIECD}          & Faith   & Decoding  & Mid  & Mid  \\
\cellcolor{Paired3} PH3 +memory    & Robust  & Mech.  & High & Low  \\
\cellcolor{Paired4} \textcolor{white}{ACD}             & Robust  & Decoding  & None & Mid  \\
\bottomrule
\end{tabular}}
\caption{Comparison of CMTs by objective, intervention level, and cost. 
`Mech.' denotes mechanistic interventions.}
\label{tab:cmts_summary}
\end{table}


%% file: Tables/pareto.tex
\begin{table}[ht]
    \centering
    \resizebox{\columnwidth}{!}{
    \begin{tabular}{lcc}
        \toprule
         & Faithfulness & Robustness \\
        \midrule
        \multicolumn{3}{c}{CounterFact} \\
        \midrule
        (Qwen 32B, Prompting)      & \textbf{100.0} & 80.67 \\
        (Pythia, Prompting)        & 99.82  & 86.07 \\
        (Pythia, Fine-tuning)      & 82.53  & 89.44 \\
        (Pythia, Regular)          & 78.27  & 91.48 \\
        (Qwen 1.5B-I, Multi-agent) & 61.64  & 99.88 \\
        (Qwen 32B-I, Multi-agent)  & 60.32  & \textbf{100.0} \\
        \midrule
        \multicolumn{3}{c}{NQ} \\
        \midrule
        (Qwen 32B, Fine-tuning)    & \textbf{74.22}  & 46.28 \\
        (Qwen 32B-I, ACD)          & 67.66  & 57.35 \\
        (Qwen 32B, ACD)            & 65.88  & 57.59 \\
        (Qwen 7B-I, Multi-agent)   & 59.14  & \textbf{73.32} \\
        \midrule
        \multicolumn{3}{c}{DRUID} \\
        \midrule
        (Qwen 32B-I, Multi-agent)  & \textbf{74.34}  & 94.12 \\
        (Qwen 1.5B, COIECD)        & 46.33  & \textbf{100.0} \\
        \bottomrule
    \end{tabular}}
    \caption{Results of the Pareto frontier analysis. LM-CMT pairs are sorted by faithfulness score. Best scores are highlighted in \textbf{bold}.}
    \label{tab:pareto}
\end{table}

%% file: Tables/featureCorrs/model_agg.tex
\definecolor{GrayRegular}{RGB}{179,179,179}

\begin{table}[ht!]
\scriptsize
\centering
\begin{tabular}{lllr}
\toprule
Dataset & Context & CMT & Corr. \\
\midrule
\multicolumn{4}{c}{\textbf{Model size}} \\
\midrule
DRUID & Gold & \cellcolor{Paired8} \textcolor{white}{Multi-agent} & \textbf{0.42} \\
DRUID & Gold & \cellcolor{Paired4} \textcolor{white}{ACD} & \textbf{0.41} \\
NQ & Gold & \cellcolor{Paired3} PH3 +memory & \textbf{0.37} \\
DRUID & Gold & \cellcolor{GrayRegular} Regular & \textbf{0.36} \\
DRUID & Gold & \cellcolor{Paired7} Prompting & \textbf{0.36} \\
NQ & Conflicting & \cellcolor{Paired3} PH3 +memory & \textbf{0.33} \\
NQ & Gold & \cellcolor{GrayRegular} Regular & \color{gray}\textbf{0.20} \\
NQ & Irrelevant & \cellcolor{GrayRegular} Regular & \color{gray}\textbf{0.14} \\
NQ & Conflicting & \cellcolor{GrayRegular} Regular & \color{gray}\textbf{0.09} \\
CounterFact & Gold & \cellcolor{GrayRegular} Regular & \color{gray}\textbf{0.04} \\
CounterFact & Irrelevant & \cellcolor{GrayRegular} Regular & \color{gray}0.02 \\
CounterFact & Conflicting & \cellcolor{GrayRegular} Regular & \color{gray}-0.01 \\
DRUID & Conflicting & \cellcolor{GrayRegular} Regular & \color{gray}\textbf{-0.08} \\
DRUID & Irrelevant & \cellcolor{GrayRegular} Regular & \color{gray}\textbf{-0.20} \\
DRUID & Irrelevant & \cellcolor{Paired3} PH3 +memory & \textbf{-0.33} \\
CounterFact & Conflicting & \cellcolor{Paired5} Fine-tuning & \textbf{-0.33} \\
DRUID & Irrelevant & \cellcolor{Paired2} \textcolor{white}{COIECD} & \textbf{-0.44} \\
\midrule
\multicolumn{4}{c}{\textbf{Instruct tuned}} \\
\midrule
DRUID & Conflicting & \cellcolor{Paired3} PH3 +memory & \textbf{0.77} \\
DRUID & Irrelevant & \cellcolor{Paired1} PH3 +context & \textbf{0.65} \\
DRUID & Conflicting & \cellcolor{Paired4} \textcolor{white}{ACD} & \textbf{0.54} \\
DRUID & Conflicting & \cellcolor{Paired7} Prompting & \textbf{0.46} \\
DRUID & Conflicting & \cellcolor{GrayRegular} Regular & \textbf{0.40} \\
DRUID & Conflicting & \cellcolor{Paired2} \textcolor{white}{COIECD} & \textbf{0.34} \\
DRUID & Irrelevant & \cellcolor{GrayRegular} Regular & \color{gray}\textbf{0.29} \\
NQ & Gold & \cellcolor{GrayRegular} Regular & \color{gray}\textbf{0.13} \\
CounterFact & Irrelevant & \cellcolor{GrayRegular} Regular & \color{gray}\textbf{0.12} \\
NQ & Irrelevant & \cellcolor{GrayRegular} Regular & \color{gray}\textbf{0.06} \\
NQ & Conflicting & \cellcolor{GrayRegular} Regular & \color{gray}\textbf{0.05} \\
CounterFact & Gold & \cellcolor{GrayRegular} Regular & \color{gray}0.01 \\
DRUID & Gold & \cellcolor{GrayRegular} Regular & \color{gray}\textbf{-0.19} \\
CounterFact & Conflicting & \cellcolor{GrayRegular} Regular & \textbf{-0.36} \\
DRUID & Gold & \cellcolor{Paired4} \textcolor{white}{ACD} & \textbf{-0.38} \\
CounterFact & Conflicting & \cellcolor{Paired1} PH3 +context & \textbf{-0.43} \\
DRUID & Gold & \cellcolor{Paired3} PH3 +memory & \textbf{-0.72} \\
\midrule
\multicolumn{4}{c}{\textbf{Strength of memory}} \\
\midrule
DRUID & Conflicting & \cellcolor{Paired3} PH3 +memory & \textbf{0.54} \\
NQ & Irrelevant & \cellcolor{Paired5} Fine-tuning & \textbf{0.47} \\
NQ & Irrelevant & \cellcolor{Paired4} \textcolor{white}{ACD} & \textbf{0.39} \\
CounterFact & Irrelevant & \cellcolor{Paired5} Fine-tuning & \textbf{0.39} \\
NQ & Irrelevant & \cellcolor{Paired7} Prompting & \textbf{0.39} \\
NQ & Irrelevant & \cellcolor{Paired2} \textcolor{white}{COIECD} & \textbf{0.38} \\
DRUID & Conflicting & \cellcolor{Paired4} \textcolor{white}{ACD} & \textbf{0.37} \\
NQ & Irrelevant & \cellcolor{GrayRegular} Regular & \textbf{0.37} \\
CounterFact & Irrelevant & \cellcolor{GrayRegular} Regular & \textbf{0.35} \\
DRUID & Conflicting & \cellcolor{Paired7} Prompting & \textbf{0.34} \\
CounterFact & Irrelevant & \cellcolor{Paired4} \textcolor{white}{ACD} & \textbf{0.32} \\
CounterFact & Irrelevant & \cellcolor{Paired3} PH3 +memory & \textbf{0.31} \\
CounterFact & Irrelevant & \cellcolor{Paired2} \textcolor{white}{COIECD} & \textbf{0.30} \\
DRUID & Conflicting & \cellcolor{GrayRegular} Regular & \color{gray}\textbf{0.26} \\
NQ & Gold & \cellcolor{GrayRegular} Regular & \color{gray}\textbf{0.18} \\
DRUID & Irrelevant & \cellcolor{GrayRegular} Regular & \color{gray}0.15 \\
NQ & Conflicting & \cellcolor{GrayRegular} Regular & \color{gray}\textbf{0.09} \\
CounterFact & Gold & \cellcolor{GrayRegular} Regular & \color{gray}\textbf{0.04} \\
DRUID & Gold & \cellcolor{GrayRegular} Regular & \color{gray}0.02 \\
CounterFact & Conflicting & \cellcolor{Paired4} \textcolor{white}{ACD} & \textbf{-0.31} \\
CounterFact & Conflicting & \cellcolor{Paired2} \textcolor{white}{COIECD} & \textbf{-0.42} \\
DRUID & Gold & \cellcolor{Paired3} PH3 +memory & \textbf{-0.43} \\
CounterFact & Conflicting & \cellcolor{GrayRegular} Regular & \textbf{-0.44} \\
\bottomrule
\end{tabular}
\caption{Spearman's $\rho$ between BCU and model features. Correlation values for \texttt{Regular} or with an absolute value above 0.3 are shown. Correlation values with an absolute value below 0.3 are marked in {\color{gray}gray}. Significant correlation values (p-value < 0.05) are marked in \textbf{bold}. Results are measured across models.}
\label{tab:corr_model}
\end{table}

%% file: Tables/featureCorrs/input_agg.tex
\begin{table}[ht!]
\scriptsize
\centering
\begin{tabular}{lllr}
\toprule
Dataset & Context & CMT & Corr. \\
\midrule
\multicolumn{4}{c}{\textbf{Context length}} \\
\midrule
CounterFact & Conflicting & \cellcolor{GrayRegular} Regular & \color{gray}\textbf{0.05} \\
CounterFact & Irrelevant & \cellcolor{GrayRegular} Regular & \color{gray}\textbf{0.03} \\
CounterFact & Gold & \cellcolor{GrayRegular} Regular & \color{gray}0.02 \\
DRUID & Conflicting & \cellcolor{GrayRegular} Regular & \color{gray}\textbf{-0.01} \\
NQ & Irrelevant & \cellcolor{GrayRegular} Regular & \color{gray}\textbf{-0.05} \\
DRUID & Irrelevant & \cellcolor{GrayRegular} Regular & \color{gray}-0.05 \\
DRUID & Gold & \cellcolor{GrayRegular} Regular & \color{gray}\textbf{-0.08} \\
NQ & Gold & \cellcolor{GrayRegular} Regular & \color{gray}\textbf{-0.20} \\
NQ & Conflicting & \cellcolor{GrayRegular} Regular & \color{gray}\textbf{-0.21} \\
DRUID & Irrelevant & \cellcolor{Paired8} \textcolor{white}{Multi-agent} & \textbf{-0.32} \\
\midrule
\multicolumn{4}{c}{\textbf{Query-context overlap}} \\
\midrule
DRUID & Irrelevant & \cellcolor{GrayRegular} Regular & \color{gray}0.03 \\
DRUID & Gold & \cellcolor{GrayRegular} Regular & \color{gray}0.01 \\
NQ & Gold & \cellcolor{GrayRegular} Regular & \color{gray}\textbf{-0.08} \\
NQ & Conflicting & \cellcolor{GrayRegular} Regular & \color{gray}\textbf{-0.08} \\
NQ & Irrelevant & \cellcolor{GrayRegular} Regular & \color{gray}\textbf{-0.10} \\
DRUID & Conflicting & \cellcolor{GrayRegular} Regular & \color{gray}\textbf{-0.13} \\
\midrule
\multicolumn{4}{c}{\textbf{Distractor rate}} \\
\midrule
CounterFact & Gold & \cellcolor{GrayRegular} Regular & \color{gray}0.02 \\
NQ & Gold & \cellcolor{GrayRegular} Regular & \color{gray}\textbf{-0.15} \\
NQ & Conflicting & \cellcolor{GrayRegular} Regular & \color{gray}\textbf{-0.16} \\
CounterFact & Conflicting & \cellcolor{GrayRegular} Regular & \color{gray}\textbf{-0.19} \\
CounterFact & Conflicting & \cellcolor{Paired4} \textcolor{white}{ACD} & \textbf{-0.34} \\
CounterFact & Conflicting & \cellcolor{Paired8} \textcolor{white}{Multi-agent} & \textbf{-0.36} \\
\midrule
\multicolumn{4}{c}{\textbf{Relevance judgement}} \\
\midrule
CounterFact & Conflicting & \cellcolor{Paired8} \textcolor{white}{Multi-agent} & \textbf{0.41} \\
CounterFact & Conflicting & \cellcolor{GrayRegular} Regular & \color{gray}\textbf{0.15} \\
NQ & Irrelevant & \cellcolor{GrayRegular} Regular & \color{gray}\textbf{0.11} \\
NQ & Gold & \cellcolor{GrayRegular} Regular & \color{gray}\textbf{0.04} \\
DRUID & Irrelevant & \cellcolor{GrayRegular} Regular & \color{gray}0.04 \\
DRUID & Gold & \cellcolor{GrayRegular} Regular & \color{gray}\textbf{0.03} \\
NQ & Conflicting & \cellcolor{GrayRegular} Regular & \color{gray}\textbf{0.02} \\
CounterFact & Irrelevant & \cellcolor{GrayRegular} Regular & \color{gray}0.01 \\
CounterFact & Gold & \cellcolor{GrayRegular} Regular & \color{gray}-0.00 \\
DRUID & Conflicting & \cellcolor{GrayRegular} Regular & \color{gray}\textbf{-0.15} \\
NQ & Irrelevant & \cellcolor{Paired8} \textcolor{white}{Multi-agent} & \textbf{-0.33} \\
DRUID & Irrelevant & \cellcolor{Paired8} \textcolor{white}{Multi-agent} & \textbf{-0.48} \\
\bottomrule
\end{tabular}
\caption{Spearman's $\rho$ between BCU and different input aspects. Correlation vallues for \texttt{Regular} or with an absolute value above 0.3 are shown. Correlation values with an absolute value below 0.3 are marked in {\color{gray}gray}. Significant correlation values (p-value < 0.05) are marked in \textbf{bold}. Results are measured across models.}
\label{tab:corr_input}
\end{table}

%% file: Tables/mainTableBCU.tex
\begin{table*}[tp]
\scriptsize
\centering
\begin{tabular}{ll|cccc|cccc|cccc}
\toprule
 & Dataset & \multicolumn{4}{c|}{CounterFact} & \multicolumn{4}{c|}{NQ} & \multicolumn{4}{c}{DRUID}  \\

 \midrule
Model & Method & Gold & Conflict. & Irrel. & Tot. & Gold & Conflict. & Irrel. & Tot. & Gold & Conflict. & Irrel. & Tot. \\

\midrule

\multirow{7}{*}{{\scshape GPT-2 XL}} & Regular & \textbf{100.0} & 96.4 & 81.0 & 92.5 & 43.0 & 37.6 & 13.7 & 31.4 & 80.9 & 7.3 & 76.5 & 39.6 \\
 & Fine-tuning & \textbf{100.0} & 92.9 & 82.4 & 91.8 & 46.9 & 42.3 & 13.9 & 34.3 & 72.4 & 12.6 & 47.1 & 38.7 \\
 & Prompting & \textbf{100.0} & 96.4 & 81.0 & 92.5 & 42.4 & 36.2 & 14.2 & 30.9 & 83.3 & 1.9 & \textbf{100.0} & 37.7 \\
 & PH3 +context & \textbf{100.0} & 99.4 & 44.8 & 81.4 & 42.3 & 36.4 & 14.0 & 30.9 & 79.6 & 11.6 & 76.5 & 41.5 \\
 & PH3 +memory & \textbf{100.0} & 99.5 & 76.8 & 92.1 & 41.4 & 35.4 & 13.9 & 30.2 & 81.1 & 3.9 & \textbf{100.0} & 37.9 \\
 & COIECD & \textbf{100.0} & 97.6 & 70.8 & 89.5 & 43.4 & 37.4 & 9.0 & 29.9 & 69.8 & 21.3 & 41.2 & 42.4 \\
 & ACD & 99.6 & 49.1 & 91.0 & 79.9 & 31.8 & 29.1 & 36.4 & 32.4 & 81.3 & 3.2 & \textbf{100.0} & 37.6 \\
\midrule 

\multirow{7}{*}{{\scshape Pythia 6.9B}} & Regular & \textbf{100.0} & 56.5 & 91.5 & 82.7 & 52.7 & 43.9 & 16.2 & 37.6 & 84.1 & 9.4 & 52.9 & 42.1 \\
 & Fine-tuning & \textbf{100.0} & 65.1 & 89.4 & 84.8 & 54.0 & 49.6 & 14.6 & 39.4 & 81.5 & 1.4 & 94.1 & 36.6 \\
 & Prompting & \textbf{100.0} & 99.6 & 86.1 & 95.2 & 52.7 & 43.9 & 16.2 & 37.6 & 82.8 & 7.1 & 64.7 & 40.3 \\
 & PH3 +context & 98.3 & 89.7 & 62.4 & 83.5 & 55.9 & 46.3 & 14.6 & 38.9 & 87.1 & 8.7 & 58.8 & 43.0 \\
 & PH3 +memory & 91.4 & 4.0 & 90.5 & 61.9 & 48.9 & 39.2 & 18.1 & 35.4 & 86.2 & 8.4 & 70.6 & 42.5 \\
 & COIECD & 99.9 & 66.0 & 86.0 & 84.0 & 53.9 & 43.8 & 10.2 & 35.9 & 72.0 & 13.0 & 41.2 & 38.8 \\
 & ACD & \textbf{100.0} & 9.7 & 96.0 & 68.6 & 43.8 & 36.1 & 32.6 & 37.5 & 87.4 & 5.2 & \textbf{100.0} & 41.3 \\
\midrule 

\multirow{7}{*}{{\scshape Qwen2.5 1.5B}} & Regular & 99.9 & 53.1 & 80.0 & 77.6 & 44.0 & 41.1 & 22.4 & 35.8 & 84.7 & 11.6 & 70.6 & 43.6 \\
 & Fine-tuning & \textbf{100.0} & 90.3 & 85.7 & 92.0 & 66.1 & 61.9 & 18.5 & 48.8 & 79.7 & 18.5 & 52.9 & 45.3 \\
 & Prompting & \textbf{100.0} & 97.2 & 82.2 & 93.2 & 63.9 & 57.5 & 32.1 & 51.1 & 85.0 & 7.0 & 82.4 & 41.2 \\
 & PH3 +context & \textbf{100.0} & 99.0 & 62.5 & 87.2 & 44.2 & 40.9 & 21.7 & 35.6 & 63.8 & 40.4 & 17.6 & 50.5 \\
 & PH3 +memory & 98.9 & 38.5 & 84.9 & 74.1 & 19.4 & 17.3 & 26.0 & 20.9 & 81.2 & 1.4 & \textbf{100.0} & 36.5 \\
 & COIECD & 94.8 & 1.2 & 89.8 & 61.9 & 42.4 & 39.2 & 45.8 & 42.5 & 87.8 & 4.8 & \textbf{100.0} & 41.3 \\
 & ACD & 97.6 & 7.7 & 90.3 & 65.2 & 46.7 & 42.8 & 39.3 & 42.9 & 87.8 & 4.8 & \textbf{100.0} & 41.3 \\
\midrule 

\multirow{8}{*}{\shortstack[l]{{\scshape Qwen2.5 1.5B} \\ \emph{{Instruct}}}} & Regular & 97.6 & 31.7 & 86.2 & 71.8 & 70.1 & 62.8 & 28.2 & 53.7 & 47.3 & 70.3 & 94.1 & 60.4 \\
 & Fine-tuning & \textbf{100.0} & 93.2 & 82.7 & 92.0 & 51.0 & 45.6 & 42.2 & 46.3 & 72.0 & 14.5 & 29.4 & 39.6 \\
 & Prompting & 99.3 & 94.2 & 76.1 & 89.9 & 68.1 & 60.5 & 29.1 & 52.5 & 47.3 & 70.3 & 94.1 & 60.4 \\
 & Multi-agent & 98.6 & 24.7 & 99.9 & 74.4 & 68.5 & 60.2 & 45.0 & 57.9 & 44.4 & 72.4 & 94.1 & 60.3 \\
 & PH3 +context & 96.0 & 42.5 & 59.8 & 66.1 & 67.1 & 59.9 & 26.0 & 51.0 & 61.1 & 64.7 & 94.1 & 63.2 \\
 & PH3 +memory & 94.6 & 11.5 & 85.5 & 63.9 & 48.8 & 42.7 & 22.0 & 37.8 & 25.4 & 76.1 & 94.1 & 54.1 \\
 & COIECD & 97.8 & 35.8 & 82.7 & 72.1 & 70.5 & 63.9 & 22.1 & 52.1 & 64.1 & 59.6 & 94.1 & 61.7 \\
 & ACD & 95.6 & 12.1 & 93.5 & 67.1 & 66.7 & 60.0 & 43.4 & 56.7 & 12.3 & 79.9 & 94.1 & 50.6 \\
\midrule 

\multirow{7}{*}{{\scshape Qwen2.5 7B}} & Regular & 96.6 & 36.0 & 79.0 & 70.5 & 71.7 & 65.6 & 25.3 & 54.2 & 91.8 & 23.6 & 41.2 & 53.3 \\
 & Fine-tuning & 99.6 & 47.4 & 85.0 & 77.4 & 76.7 & 68.8 & 41.7 & 62.4 & 86.4 & 1.8 & 82.4 & 39.0 \\
 & Prompting & \textbf{100.0} & 97.8 & 81.3 & 93.0 & 74.7 & 66.5 & 31.2 & 57.5 & 94.9 & 13.8 & 58.8 & 49.3 \\
 & PH3 +context & 97.8 & 96.3 & 16.7 & 70.3 & 69.7 & 63.6 & 25.3 & 52.8 & 83.4 & 50.1 & 17.6 & 64.5 \\
 & PH3 +memory & 96.8 & 4.0 & 84.2 & 61.6 & 66.5 & 59.5 & 26.6 & 50.8 & 90.5 & 4.1 & 76.5 & 42.0 \\
 & COIECD & 96.6 & 36.0 & 79.0 & 70.5 & 71.7 & 65.6 & 25.3 & 54.2 & 91.8 & 23.6 & 41.2 & 53.3 \\
 & ACD & 94.7 & 2.3 & 92.7 & 63.2 & 72.3 & 59.9 & 41.9 & 58.0 & 89.8 & 12.6 & 70.6 & 46.4 \\
\midrule 

\multirow{8}{*}{\shortstack[l]{{\scshape Qwen2.5 7B} \\ \emph{{Instruct}}}} & Regular & \textbf{100.0} & 25.9 & 84.5 & 70.1 & 76.2 & 65.0 & 31.0 & 57.4 & 87.8 & 57.1 & 64.7 & 70.5 \\
 & Fine-tuning & \textbf{100.0} & 62.3 & 81.0 & 81.1 & 59.6 & 52.7 & 48.1 & 53.5 & 96.4 & 13.2 & 70.6 & 49.6 \\
 & Prompting & \textbf{100.0} & 98.6 & 35.3 & 78.0 & 75.8 & 66.7 & 29.1 & 57.2 & 87.8 & 57.1 & 64.7 & 70.5 \\
 & Multi-agent & 95.7 & 11.6 & \textbf{100.0} & 69.1 & 66.1 & 52.2 & 73.3 & 63.9 & 58.6 & 63.2 & 94.1 & 61.3 \\
 & PH3 +context & 98.3 & 84.0 & 54.1 & 78.8 & 75.3 & 64.4 & 26.9 & 55.5 & 86.9 & 54.7 & 70.6 & 68.8 \\
 & PH3 +memory & \textbf{100.0} & 27.6 & 82.8 & 70.1 & 76.4 & 66.1 & 30.9 & 57.8 & 3.1 & \textbf{81.4} & 70.6 & 47.3 \\
 & COIECD & 99.9 & 9.1 & 90.6 & 66.5 & 76.2 & 60.1 & 40.8 & 59.0 & 76.4 & 56.5 & 76.5 & 65.2 \\
 & ACD & 99.6 & 11.5 & 96.9 & 69.3 & 76.3 & 62.1 & 44.6 & 61.0 & 76.2 & 57.6 & 76.5 & 65.8 \\
\midrule 

\multirow{5}{*}{{\scshape Qwen2.5 32B}} & Regular & 99.9 & 77.6 & 77.2 & 84.9 & 77.3 & 66.7 & 39.7 & 61.2 & \textbf{98.2} & 19.8 & 41.2 & 54.0 \\
 & Fine-tuning & 98.1 & 88.4 & 81.9 & 89.4 & 79.2 & \textbf{69.2} & 46.3 & 64.9 & 98.0 & 9.7 & 82.4 & 48.4 \\
 & Prompting & \textbf{100.0} & \textbf{100.0} & 80.7 & 93.6 & 77.2 & 66.9 & 42.8 & 62.3 & \textbf{98.2} & 22.5 & 52.9 & 55.6 \\
 & COIECD & 97.4 & 96.5 & 58.5 & 84.1 & 76.1 & 67.4 & 32.7 & 58.7 & 97.1 & 27.8 & 29.4 & 57.9 \\
 & ACD & 97.6 & 2.3 & 92.6 & 64.1 & 75.7 & 56.1 & 57.6 & 63.1 & 97.6 & 14.1 & 58.8 & 50.6 \\
\midrule 

\multirow{6}{*}{\shortstack[l]{{\scshape Qwen2.5 32B} \\ \emph{{Instruct}}}} & Regular & 99.4 & 4.9 & 92.6 & 65.6 & 81.4 & 59.9 & 43.8 & 61.7 & 97.9 & 43.2 & 76.5 & 67.2 \\
 & Fine-tuning & \textbf{100.0} & 18.0 & 93.6 & 70.5 & 71.6 & 64.9 & 42.0 & 59.5 & 96.4 & 20.8 & 52.9 & 53.8 \\
 & Prompting & 99.9 & 95.3 & 69.1 & 88.1 & 81.4 & 59.9 & 43.8 & 61.7 & 97.2 & 48.7 & 82.4 & 70.0 \\
 & Multi-agent & \textbf{100.0} & 20.6 & \textbf{100.0} & 73.5 & 76.8 & 57.2 & 49.2 & 61.1 & 93.1 & 55.6 & 94.1 & 72.1 \\
 & COIECD & 98.0 & 6.0 & 70.8 & 58.3 & 79.7 & 61.6 & 36.8 & 59.4 & 97.7 & 38.3 & 64.7 & 64.3 \\
 & ACD & 98.4 & 2.5 & 97.5 & 66.1 & 80.1 & 55.2 & 57.4 & 64.2 & 88.5 & 51.4 & 94.1 & 67.7 \\
\midrule 

\multirow{3}{*}{{\scshape Command A}} & Regular & \textbf{100.0} & \textbf{100.0} & 4.1 & 68.0 & 79.2 & 62.7 & 28.9 & 56.9 & 95.9 & 57.3 & 76.5 & 74.2 \\
 & Prompting & 97.0 & 92.8 & 48.4 & 79.4 & 79.2 & 62.7 & 28.9 & 56.9 & 93.6 & 64.4 & 70.6 & \textbf{77.2} \\
 & Multi-agent & 99.6 & 39.1 & 99.9 & 79.6 & 74.3 & 49.7 & 58.8 & 61.0 & 91.9 & 48.2 & 94.1 & 67.4 \\
\midrule 

 \multirow{3}{*}{{\scshape GPT-4.1 mini}} & Regular & 99.4 & 2.6 & 92.4 & 64.8 & 80.2 & 54.7 & 47.3 & 60.8 & 96.6 & 51.8 & 58.8 & 71.4 \\
 & Prompting & \textbf{100.0} & \textbf{100.0} & 78.3 & 92.8 & 76.7 & 47.0 & 51.0 & 58.3 & 82.9 & 44.7 & 35.3 & 61.3 \\
 & Multi-agent & 99.6 & 55.6 & \textbf{100.0} & 85.1 & 77.7 & 48.0 & 68.4 & 64.8 & 96.9 & 47.6 & \textbf{100.0} & 69.2 \\
\midrule 

\multirow{3}{*}{{\scshape GPT-4.1}} & Regular & 99.3 & 25.7 & 96.0 & 73.7 & \textbf{84.2} & 54.2 & 52.2 & 63.5 & 97.5 & 47.0 & 41.2 & 68.9 \\
 & Prompting & \textbf{100.0} & 95.6 & 94.0 & \textbf{96.5} & 80.7 & 49.2 & 56.0 & 62.0 & 96.4 & 60.7 & 35.3 & 76.1 \\
 & Multi-agent & 99.8 & 63.4 & \textbf{100.0} & 87.7 & 82.0 & 45.7 & \textbf{76.0} & \textbf{68.0} & 97.7 & 37.5 & 94.1 & 63.9 \\
 
\bottomrule
\end{tabular}
\caption{$\mathrm{BCU}$ scores on CUB. A high $\mathrm{BCU}$ score is desirable regardless of context type. Gold denotes relevant contexts that also contain the gold answer. Conflict. denotes `Conflicting' -- relevant contexts that contain a conflicting answer, dissimilar from the correct answer or model memory. Irrel. denotes irrelevant contexts. Tot. denotes the average performance across all context types. Values marked in \textbf{bold} indicate the top CMT score across LMs for each dataset and context type.}
\label{tab:main_results_bcu}
\end{table*}

%% file: Tables/mainTableAcc.tex
\begin{table*}[tp]
\scriptsize
\centering
\begin{tabular}{ll|cccc|cccc|cccc}
\toprule
 & Dataset & \multicolumn{4}{c|}{CounterFact} & \multicolumn{4}{c|}{NQ} & \multicolumn{4}{c}{DRUID}  \\

 \midrule
Model & Method & Gold & Conflict. & Irrel. & Tot. & Gold & Conflict. & Irrel. & Tot. & Gold & Conflict. & Irrel. & Tot. \\

\midrule

\multirow{7}{*}{{\scshape GPT-2 XL}} & Regular & \textbf{100.0} & 2.9 & 69.7 & 57.5 & 43.0 & 8.1 & 20.8 & 24.0 & 80.9 & 69.0 & 64.7 & 74.2 \\
 & Fine-tuning & \textbf{100.0} & 3.2 & 70.6 & 57.9 & 46.9 & 7.7 & 23.8 & 26.2 & 72.4 & 65.5 & 41.2 & 68.4 \\
 & Prompting & \textbf{100.0} & 2.9 & 69.7 & 57.5 & 42.4 & 7.5 & 20.3 & 23.5 & 83.3 & 73.8 & 76.5 & 78.0 \\
 & PH3 +context & \textbf{100.0} & 0.4 & 29.8 & 43.4 & 42.3 & 7.8 & 20.4 & 23.6 & 79.6 & 65.7 & 52.9 & 71.7 \\
 & PH3 +memory & \textbf{100.0} & 0.4 & 65.1 & 55.1 & 41.4 & 7.4 & 20.1 & 23.0 & 81.1 & 72.6 & 76.5 & 76.3 \\
 & COIECD & \textbf{100.0} & 2.3 & 67.7 & 56.7 & 43.4 & 7.1 & 19.4 & 23.3 & 69.8 & 51.0 & 47.1 & 59.1 \\
 & ACD & 99.6 & 29.4 & 72.3 & 67.1 & 31.8 & 7.7 & 18.1 & 19.2 & 81.3 & 73.0 & 76.5 & 76.6 \\
\midrule 

\multirow{7}{*}{{\scshape Pythia 6.9B}} & Regular & \textbf{100.0} & 37.2 & 91.4 & 76.2 & 52.7 & 9.8 & 29.6 & 30.8 & 84.1 & 49.9 & 47.1 & 64.7 \\
 & Fine-tuning & \textbf{100.0} & 26.5 & 91.8 & 72.8 & 54.0 & 5.6 & 26.6 & 28.8 & 81.5 & 74.4 & 70.6 & 77.5 \\
 & Prompting & \textbf{100.0} & 0.5 & 86.1 & 62.2 & 52.7 & 9.8 & 29.6 & 30.8 & 82.8 & 57.1 & 47.1 & 68.3 \\
 & PH3 +context & 98.3 & 2.5 & 62.1 & 54.3 & 55.9 & 8.4 & 30.0 & 31.5 & 87.1 & 55.2 & 52.9 & 69.0 \\
 & PH3 +memory & 91.4 & 86.0 & 90.4 & 89.2 & 48.9 & 11.5 & 29.7 & 30.1 & 86.2 & 55.1 & 64.7 & 68.7 \\
 & COIECD & 99.9 & 27.3 & 86.0 & 71.0 & 53.9 & 9.8 & 27.4 & 30.4 & 72.0 & 32.9 & 35.3 & 50.0 \\
 & ACD & \textbf{100.0} & 77.6 & 95.9 & 91.2 & 43.8 & 12.1 & 29.7 & 28.6 & 87.4 & 69.2 & \textbf{82.4} & 77.2 \\
\midrule 

\multirow{7}{*}{{\scshape Qwen2.5 1.5B}} & Regular & 99.9 & 41.9 & 74.2 & 72.0 & 44.0 & 7.7 & 22.0 & 24.6 & 84.7 & 63.5 & 52.9 & 72.7 \\
 & Fine-tuning & \textbf{100.0} & 5.5 & 77.0 & 60.8 & 66.1 & 18.8 & 42.4 & 42.5 & 79.7 & 60.3 & 58.8 & 68.7 \\
 & Prompting & \textbf{100.0} & 1.6 & 79.7 & 60.4 & 63.9 & 17.0 & 38.5 & 39.8 & 85.0 & 69.8 & 58.8 & 76.4 \\
 & PH3 +context & \textbf{100.0} & 0.7 & 50.1 & 50.3 & 44.2 & 12.6 & 25.5 & 27.5 & 63.8 & 26.9 & 11.8 & 42.9 \\
 & PH3 +memory & 98.9 & 52.8 & 78.0 & 76.6 & 19.4 & 8.1 & 10.4 & 12.7 & 81.2 & 74.5 & 70.6 & 77.4 \\
 & COIECD & 94.8 & 71.9 & 79.0 & 81.9 & 42.4 & 16.3 & 27.6 & 28.8 & 87.8 & 72.7 & 70.6 & 79.3 \\
 & ACD & 97.6 & 70.8 & 79.4 & 82.6 & 46.7 & 15.5 & 28.0 & 30.1 & 87.8 & 72.7 & 70.6 & 79.3 \\
\midrule 

\multirow{8}{*}{\shortstack[l]{{\scshape Qwen2.5 1.5B} \\ \emph{{Instruct}}}} & Regular & 97.6 & 54.5 & 79.6 & 77.2 & 70.1 & 16.1 & 37.1 & 41.2 & 47.3 & 11.1 & 0.0 & 26.8 \\
 & Fine-tuning & \textbf{100.0} & 7.0 & 78.0 & 61.7 & 51.0 & 7.6 & 27.8 & 28.8 & 72.0 & 28.5 & 47.1 & 47.5 \\
 & Prompting & 99.3 & 5.4 & 74.1 & 59.6 & 68.1 & 15.7 & 38.8 & 41.0 & 47.3 & 11.1 & 0.0 & 26.8 \\
 & Multi-agent & 98.6 & 68.7 & 83.0 & 83.4 & 68.5 & 16.9 & 36.1 & 40.6 & 44.4 & 10.0 & 0.0 & 24.9 \\
 & PH3 +context & 96.0 & 35.9 & 58.2 & 63.4 & 67.1 & 15.4 & 34.7 & 39.1 & 61.1 & 18.9 & 0.0 & 37.2 \\
 & PH3 +memory & 94.6 & 68.9 & 78.3 & 80.6 & 48.8 & 13.1 & 25.8 & 29.3 & 25.4 & 7.2 & 0.0 & 15.1 \\
 & COIECD & 97.8 & 50.4 & 77.1 & 75.1 & 70.5 & 15.5 & 35.9 & 40.7 & 64.1 & 19.2 & 0.0 & 38.7 \\
 & ACD & 95.6 & 77.7 & 82.1 & 85.1 & 66.7 & 19.0 & 39.0 & 41.6 & 12.3 & 3.6 & 0.0 & 7.4 \\
\midrule 

\multirow{7}{*}{{\scshape Qwen2.5 7B}} & Regular & 96.6 & 52.2 & 72.6 & 73.8 & 71.7 & 16.7 & 39.0 & 42.6 & 91.8 & 57.6 & 23.5 & 72.3 \\
 & Fine-tuning & 99.6 & 45.1 & 77.1 & 73.9 & 76.7 & 18.5 & 50.5 & 48.6 & 86.4 & \textbf{74.8} & 70.6 & 79.8 \\
 & Prompting & \textbf{100.0} & 2.4 & 86.2 & 62.9 & 74.7 & 17.9 & 44.6 & 45.8 & 94.9 & 64.2 & 35.3 & 77.4 \\
 & PH3 +context & 97.8 & 0.2 & 6.0 & 34.7 & 69.7 & 17.0 & 38.7 & 41.9 & 83.4 & 30.5 & 5.9 & 53.4 \\
 & PH3 +memory & 96.8 & 88.6 & 79.4 & 88.2 & 66.5 & 17.6 & 37.7 & 40.6 & 90.5 & 73.4 & 70.6 & \textbf{80.8} \\
 & COIECD & 96.6 & 52.2 & 72.6 & 73.8 & 71.7 & 16.7 & 39.0 & 42.6 & 91.8 & 57.6 & 23.5 & 72.3 \\
 & ACD & 94.7 & 85.5 & 80.4 & 86.9 & 72.3 & 23.9 & 47.2 & 47.8 & 89.8 & 68.1 & 47.1 & 77.5 \\
\midrule 

\multirow{8}{*}{\shortstack[l]{{\scshape Qwen2.5 7B} \\ \emph{{Instruct}}}} & Regular & \textbf{100.0} & 42.0 & 85.4 & 75.8 & 76.2 & 19.8 & 47.1 & 47.8 & 87.8 & 28.3 & 0.0 & 54.1 \\
 & Fine-tuning & \textbf{100.0} & 34.8 & 88.0 & 74.3 & 59.6 & 8.1 & 35.3 & 34.4 & 96.4 & 65.0 & 64.7 & 78.6 \\
 & Prompting & \textbf{100.0} & 1.9 & 37.5 & 46.5 & 75.8 & 20.3 & 46.0 & 47.4 & 87.8 & 28.3 & 0.0 & 54.1 \\
 & Multi-agent & 95.7 & 85.5 & 94.0 & 91.7 & 66.1 & 21.4 & 40.9 & 42.9 & 58.6 & 18.5 & 29.4 & 36.0 \\
 & PH3 +context & 98.3 & 12.5 & 55.6 & 55.5 & 75.3 & 18.5 & 44.1 & 46.0 & 86.9 & 31.5 & 0.0 & 55.5 \\
 & PH3 +memory & \textbf{100.0} & 50.9 & 83.8 & 78.2 & 76.4 & 20.1 & 47.7 & 48.1 & 3.1 & 2.5 & 0.0 & 2.7 \\
 & COIECD & 99.9 & 75.0 & 90.8 & 88.6 & 76.2 & 25.8 & 48.2 & 50.1 & 76.4 & 29.2 & 5.9 & 49.7 \\
 & ACD & 99.6 & 85.1 & 94.0 & 92.9 & 76.3 & 25.0 & 49.3 & 50.3 & 76.2 & 29.1 & 5.9 & 49.5 \\
\midrule 

\multirow{5}{*}{{\scshape Qwen2.5 32B}} & Regular & 99.9 & 21.4 & 75.0 & 65.4 & 77.3 & 20.8 & 47.7 & 48.7 & \textbf{98.2} & 58.5 & 29.4 & 75.7 \\
 & Fine-tuning & 98.1 & 9.8 & 77.2 & 61.7 & 79.2 & 20.3 & 55.9 & 51.9 & 98.0 & 66.6 & 64.7 & 80.3 \\
 & Prompting & \textbf{100.0} & 0.2 & 80.7 & 60.3 & 77.2 & 19.9 & 50.2 & 49.2 & \textbf{98.2} & 57.5 & 41.2 & 75.2 \\
 & COIECD & 97.4 & 3.2 & 59.7 & 53.4 & 76.1 & 18.8 & 43.9 & 46.3 & 97.1 & 47.4 & 17.6 & 68.9 \\
 & ACD & 97.6 & 85.7 & 81.3 & 88.2 & 75.7 & 31.4 & 53.3 & 53.5 & 97.6 & 66.1 & 47.1 & 79.8 \\
\midrule 

\multirow{6}{*}{\shortstack[l]{{\scshape Qwen2.5 32B} \\ \emph{{Instruct}}}} & Regular & 99.4 & 81.0 & 93.5 & 91.3 & 81.4 & 28.6 & 52.2 & 54.2 & 97.9 & 41.8 & 29.4 & 66.2 \\
 & Fine-tuning & \textbf{100.0} & 78.5 & 92.2 & 90.2 & 71.6 & 13.3 & 44.3 & 43.2 & 96.4 & 61.8 & 47.1 & 76.8 \\
 & Prompting & 99.9 & 3.2 & 70.6 & 57.9 & 81.4 & 28.6 & 52.2 & 54.2 & 97.2 & 36.2 & 11.8 & 62.6 \\
 & Multi-agent & \textbf{100.0} & 78.5 & 94.7 & 91.1 & 76.8 & 22.7 & 40.7 & 46.8 & 93.1 & 31.7 & 17.6 & 58.4 \\
 & COIECD & 98.0 & 9.7 & 72.4 & 60.0 & 79.7 & 23.4 & 49.4 & 50.9 & 97.7 & 43.3 & 29.4 & 66.9 \\
 & ACD & 98.4 & 94.7 & 95.4 & \textbf{96.2} & 80.1 & \textbf{35.3} & 55.4 & \textbf{57.0} & 88.5 & 36.0 & 17.6 & 58.8 \\
\midrule 

\multirow{3}{*}{{\scshape Command A}} & Regular & \textbf{100.0} & 0.0 & 4.4 & 34.8 & 79.2 & 12.3 & 33.8 & 41.9 & 95.9 & 30.3 & 5.9 & 58.8 \\
 & Prompting & 97.0 & 0.7 & 47.8 & 48.5 & 79.2 & 12.3 & 33.8 & 41.9 & 93.6 & 23.3 & 0.0 & 53.8 \\
 & Multi-agent & 99.6 & 32.2 & 90.2 & 74.0 & 74.3 & 13.5 & 40.4 & 42.8 & 91.9 & 33.2 & 23.5 & 58.7 \\
 \midrule 

\multirow{3}{*}{{\scshape GPT-4.1 mini}} & Regular & 99.4 & \textbf{95.3} & 93.8 & 96.2 & 80.2 & 20.9 & 48.9 & 50.1 & 96.6 & 32.2 & 17.6 & 60.2 \\
 & Prompting & \textbf{100.0} & 0.0 & 81.9 & 60.6 & 76.7 & 23.6 & 49.5 & 50.0 & 82.9 & 14.7 & 0.0 & 44.3 \\
 & Multi-agent & 99.6 & 43.1 & 90.3 & 77.7 & 77.7 & 21.4 & 45.3 & 48.2 & 96.9 & 34.4 & 35.3 & 61.6 \\
\midrule 

\multirow{3}{*}{{\scshape GPT-4.1}} & Regular & 99.3 & 72.7 & 96.9 & 89.6 & \textbf{84.2} & 25.3 & 53.9 & 54.5 & 97.5 & 39.2 & 11.8 & 64.5 \\
 & Prompting & \textbf{100.0} & 4.4 & 94.7 & 66.4 & 80.7 & 26.3 & 54.7 & 54.0 & 96.4 & 26.0 & 5.9 & 56.6 \\
 & Multi-agent & 99.8 & 35.4 & \textbf{98.0} & 77.7 & 82.0 & 28.4 & \textbf{56.7} & 55.8 & 97.7 & 45.7 & 64.7 & 68.4 \\

\bottomrule
\end{tabular}
\caption{Accuracy with respect to gold label on CUB. Gold denotes relevant contexts that also contain the gold answer. Conflict. denotes `Conflicting' -- relevant contexts that contain a conflicting answer, dissimilar from the correct answer or model memory. Irrel. denotes irrelevant contexts. Tot. denotes the average performance across all context types. Values marked in \textbf{bold} indicate the top CMT score across LMs on each dataset and context type.}
\label{tab:main_results_acc}
\end{table*}

%% file: Tables/multiAgentRelevance.tex
\begin{table}[t]
\scriptsize
\centering
\resizebox{\columnwidth}{!}{
\begin{tabular}{rcccc}
\toprule
 & Gold & Conflict. & Irrel. & All \\
\midrule
\multicolumn{5}{c}{\scshape Qwen2.5 1.5B-I} \\
\midrule
CounterFact & 98.56 & 24.25 & 99.88 & 74.23 \\
NQ & 92.44 & 91.89 & 26.26 & 70.13 \\
DRUID & 93.27 & 96.52 & 17.65 & 94.79 \\

\midrule
\multicolumn{5}{c}{\scshape Qwen2.5 7B-I} \\
\midrule
CounterFact & 99.16 & 10.68 & 99.88 & 69.91 \\
NQ & 80.70 & 76.14 & 59.35 & 72.05 \\
DRUID & 82.53 & 65.56 & 94.12 & 73.06 \\
\midrule

\multicolumn{5}{c}{\scshape Qwen2.5 32B-I} \\
\midrule

CounterFact & 99.64 & 19.57 & 99.40 & 72.87 \\
NQ & 94.74 & 92.50 & 25.77 & 70.94 \\
DRUID & 98.66 & 76.25 & 88.24 & 86.05 \\
\midrule

\multicolumn{5}{c}{\scshape Command A} \\
\midrule

CounterFact & 100.00 & 99.88 & 99.88 & 99.92 \\
NQ & 94.31 & 91.82 & 37.69 & 74.56 \\
DRUID & 93.11 & 68.55 & 88.24 & 79.31 \\

\midrule

\multicolumn{5}{c}{\scshape GPT-4.1 mini} \\
\midrule

CounterFact & 99.04 & 55.58 & 100.00 & 84.87 \\
NQ & 94.25 & 79.68 & 43.50 & 72.46 \\
DRUID & 97.60 & 77.12 & 94.12 & 86.10 \\

\midrule

\multicolumn{5}{c}{\scshape GPT-4.1} \\
\midrule

CounterFact & 99.52 & 82.11 & 100.00 & 93.88 \\
NQ & 95.22 & 83.10 & 47.49 & 75.25 \\
DRUID & 99.25 & 74.97 & 88.24 & 85.59 \\

\bottomrule
\end{tabular}}
\caption{Multi-agent relevance assessment accuracy across different context types and datasets.}
\label{tab:multi_agent_rel_acc}
\end{table}

%% file: Tables/case_by_case/counterfact.tex
\begin{table*}[ht]
\centering
\scriptsize
\begin{tabular}{lllc c r r r r}
\toprule
CMT &
Prediction &
\shortstack[l]{Prediction \\ w/ Context} &
BCU &
CCU &
\shortstack[l]{Prediction \\ Prob.} &
\shortstack[l]{Prediction Prob. \\ w/ Context} &
\shortstack[l]{Memory Token Prob. \\ w/ Context} &
\shortstack[l]{Context Token Prob. \\ w/ Context} \\
\midrule
Regular        & Detroit & Detroit & True & 0.99 & 0.53 & 0.99 & 0.99 & 0.99 \\
Prompting     & Detroit & Detroit & True & 0.99 & 0.53 & 0.99 & 0.99 & 0.99 \\
Fine-tuning   & Detroit & Detroit & True & 0.85 & 0.52 & 0.93 & 0.93 & 0.93 \\
Multi-Agent   & Detroit & Detroit & True & 0.99 & 0.53 & 0.99 & 0.99 & 0.99 \\
ACD           & Detroit & Detroit & True & 0.99 & 0.53 & 0.99 & 0.99 & 0.99 \\
COIECD        & Detroit & Detroit & True & 0.98 & 0.53 & 0.99 & 0.99 & 0.99 \\
PH3 +context  & Detroit & Detroit & True & 0.99 & 0.53 & 0.99 & 0.99 & 0.99 \\
PH3 +memory   & Detroit & Detroit & True & 0.99 & 0.53 & 0.99 & 0.99 & 0.99 \\
\bottomrule
\end{tabular}
\caption{\textbf{Prompt:} ``The White Stripes originated in''. \textbf{Context (gold):} ``Fact: The White Stripes originated in Detroit.''. Corresponding model predictions and associated probabilities are shown across different CMT methods.}
\label{tab:case-counterfact-gold}
\end{table*}
\begin{table*}[ht]
\centering
\scriptsize
\begin{tabular}{lllc c r r r r}
\toprule
CMT &
Prediction &
\shortstack[l]{Prediction \\ w/ Context} &
BCU &
CCU &
\shortstack[l]{Prediction \\ Prob.} &
\shortstack[l]{Prediction Prob. \\ w/ Context} &
\shortstack[l]{Memory Token Prob. \\ w/ Context} &
\shortstack[l]{Context Token Prob. \\ w/ Context} \\
\midrule
Regular        & Detroit & B        & True  & 0.99 & 0.53 & 0.99 & 0    & 0.99 \\
Prompting     & Detroit & B        & True  & 1.00 & 0.53 & 1.00 & 0    & 1.00 \\
Fine-tuning   & Detroit & B        & True  & 0.54 & 0.52 & 0.58 & 0    & 0.58 \\
Multi-Agent   & Detroit & B        & True  & 0.99 & 0.53 & 0.99 & 1.00 & 0.99 \\
ACD           & Detroit & B        & True  & 0.99 & 0.53 & 0.99 & 0    & 0.99 \\
COIECD        & Detroit & Michigan & False & 0.19 & 0.53 & 0.54 & 0.21 & 0.19 \\
PH3 +context  & Detroit & B        & True  & 0.99 & 0.53 & 0.99 & 0    & 0.99 \\
PH3 +memory   & Detroit & B        & True  & 0.99 & 0.53 & 0.99 & 0    & 0.99 \\
\bottomrule
\end{tabular}
\caption{\textbf{Prompt:} ``The White Stripes originated in''. \textbf{Context (conflicting):} ``Fact: The White Stripes originated in Bristol.''. Corresponding model predictions and associated probabilities are shown across different CMT methods.}
\label{tab:case-counterfact-conflicting}
\end{table*}
\begin{table*}[ht!]
\centering
\scriptsize
\begin{tabular}{lllc c r r r r}
\toprule
CMT &
Prediction &
\shortstack[l]{Prediction \\ w/ Context} &
BCU &
CCU &
\shortstack[l]{Prediction \\ Prob.} &
\shortstack[l]{Prediction Prob. \\ w/ Context} &
\shortstack[l]{Memory Token Prob. \\ w/ Context} &
\shortstack[l]{Context Token Prob. \\ w/ Context} \\
\midrule
Regular        & Detroit & Michigan & False & -0.97 & 0.53 & 0.98 & 0.01 & - \\
Prompting     & Detroit & The      & False & -0.94 & 0.53 & 0.56 & 0.02 & - \\
Fine-tuning   & Detroit & Detroit  & True  & -0.97 & 0.52 & 0.26 & 0.01 & - \\
Multi-Agent   & Detroit & Detroit  & True  & -0.97 & 0.53 & 0.98 & 0.01 & - \\
ACD           & Detroit & Michigan & False & -0.96 & 0.53 & 0.98 & 0.01 & - \\
COIECD        & Detroit & Michigan & False & -0.77 & 0.53 & 0.86 & 0.12 & - \\
PH3 +context  & Detroit & C        & False & -0.99 & 0.53 & 0.94 & 0.00 & - \\
PH3 +memory   & Detroit & Michigan & False & -0.97 & 0.53 & 0.98 & 0.01 & - \\
\bottomrule
\end{tabular}
\caption{\textbf{Prompt:} ``The White Stripes originated in''. \textbf{Context (irrelevant):} ``Fact: Tsubasa: Reservoir Chronicle was created in Denmark.''. Corresponding model predictions and associated probabilities are shown across different CMT methods. 'Context Token Prob.' is not reported since there is no context token to measure this on for irrelevant context.}
\label{tab:case-counterfact-irrelevant}
\end{table*}

%% file: Tables/case_by_case/nq.tex
\begin{table*}[ht]
\centering
\scriptsize
\begin{tabular}{lllc c r r r r}
\toprule
CMT &
Prediction &
\shortstack[l]{Prediction \\ w/ Context} &
BCU &
CCU &
\shortstack[l]{Prediction \\ Prob.} &
\shortstack[l]{Prediction Prob. \\ w/ Context} &
\shortstack[l]{Memory Token Prob. \\ w/ Context} &
\shortstack[l]{Context Token Prob. \\ w/ Context} \\
\midrule
Regular        & Canada     & British & True & 0.99 & 0.40 & 0.99 & 0    & 0.99 \\
Prompting     & Canada     & British & True & 0.99 & 0.40 & 0.99 & 0    & 0.99 \\
Fine-tuning   & Vancouver & British & True & 0.90 & 0.07 & 0.90 & 0    & 0.90 \\
Multi-Agent   & Canada     & British & True & 0.99 & 0.40 & 0.99 & 0    & 0.99 \\
ACD           & Canada     & British & True & 0.99 & 0.40 & 0.99 & 0    & 0.99 \\
COIECD        & Canada     & British & True & 0.99 & 0.40 & 0.99 & 0    & 0.99 \\
PH3 +context  & Canada     & British & True & 0.99 & 0.40 & 0.99 & 0    & 0.99 \\
PH3 +memory   & Canada     & British & True & 0.99 & 0.40 & 0.99 & 0    & 0.99 \\
\bottomrule
\end{tabular}
\caption{\textbf{Prompt:} ``where's the tv show the crossing filmed?''. \textbf{Context (gold):} ``The Crossing is an American science fiction thriller series that airs on ABC and CTV . The series debuted on April 2, 2018 . On March 20, 2018, ABC released the pilot episode on their website . The series is filmed in British Columbia, Canada .''. Corresponding model predictions and associated probabilities are shown across different CMT methods.}
\label{tab:case-nq-gold}
\end{table*}
\begin{table*}[ht]
\centering
\scriptsize
\begin{tabular}{lllc c r r r r}
\toprule
CMT &
Prediction &
\shortstack[l]{Prediction \\ w/ Context} &
BCU &
CCU &
\shortstack[l]{Prediction \\ Prob.} &
\shortstack[l]{Prediction Prob. \\ w/ Context} &
\shortstack[l]{Memory Token Prob. \\ w/ Context} &
\shortstack[l]{Context Token Prob. \\ w/ Context} \\
\midrule
Regular        & Canada     & New & True & 0.99 & 0.40 & 0.99 & 0    & 0.99 \\
Prompting     & Canada     & New & True & 0.99 & 0.40 & 0.99 & 0    & 0.99 \\
Fine-tuning   & Vancouver & New & True & 0.93 & 0.07 & 0.93 & 0    & 0.93 \\
Multi-Agent   & Canada     & New & True & 0.99 & 0.40 & 0.99 & 0    & 0.99 \\
ACD           & Canada     & New & True & 0.99 & 0.40 & 0.99 & 0    & 0.99 \\
COIECD        & Canada     & New & True & 0.99 & 0.40 & 0.99 & 0    & 0.99 \\
PH3 +context  & Canada     & New & True & 0.99 & 0.40 & 0.99 & 0    & 0.99 \\
PH3 +memory   & Canada     & New & True & 0.99 & 0.40 & 0.99 & 0    & 0.99 \\
\bottomrule
\end{tabular}
\caption{\textbf{Prompt:} ``where's the tv show the crossing filmed?''. \textbf{Context (conflicting):} ``The Crossing is an American science fiction thriller series that airs on ABC and CTV . The series debuted on April 2, 2018 . On March 20, 2018, ABC released the pilot episode on their website . The series is filmed in New York City .''. Corresponding model predictions and associated probabilities are shown across different CMT methods.}
\label{tab:case-nq-conflicting}
\end{table*}
\begin{table*}[ht!]
\centering
\scriptsize
\begin{tabular}{lllc c r r r r}
\toprule
CMT &
Prediction &
\shortstack[l]{Prediction \\ w/ Context} &
BCU &
CCU &
\shortstack[l]{Prediction \\ Prob.} &
\shortstack[l]{Prediction Prob. \\ w/ Context} &
\shortstack[l]{Memory Token Prob. \\ w/ Context} &
\shortstack[l]{Context Token Prob. \\ w/ Context} \\
\midrule
Regular        & Canada    & Information & False & -0.99 & 0.40 & 0.92 & 0   & - \\
Prompting     & Canada    & Information & False & -0.99 & 0.40 & 0.82 & 0    & - \\
Fine-tuning   & Vancouver & ABC         & False & -0.50 & 0.07 & 0.11 & 0.03 & - \\
Multi-Agent   & Canada    & Canada      & True  & -0.99 & 0.40 & 0.92 & 0    & - \\
ACD           & Canada    & Information & False & -0.99 & 0.40 & 0.57 & 0    & - \\
COIECD        & Canada    & The         & False & -0.26 & 0.40 & 0.43 & 0.29 & - \\
PH3 +context  & Canada    & Information & False & -0.99 & 0.40 & 0.86 & 0    & - \\
PH3 +memory   & Canada    & Information & False & -0.99 & 0.40 & 0.83 & 0    & - \\
\bottomrule
\end{tabular}
\caption{\textbf{Prompt:} ``where's the tv show the crossing filmed?''. \textbf{Context (irrelevant):} ``<Table> <Tr> <Th colspan="2"> The Crossing </Th> </Tr> <Tr> <Td colspan="2"> </Td> </Tr> <Tr> <Th> Created by </Th> <Td> <Ul> <Li> Jay Beattie </Li> <Li> Dan Dworkin </Li> </Ul> </Td> </Tr> <Tr> <Th> Starring </Th> <Td> <Ul> <Li> Steve Zahn </Li> <Li> Natalie Martinez </Li> <Li> Sandrine Holt </Li> </Ul> </Td> </Tr> <Tr> <Th> Opening theme </Th> <Td> "Letting the Cables Sleep" by Bush </Td> </Tr> <Tr> <Th> Composer (s) </Th> <Td> Robert Duncan </Td> </Tr> <Tr> <Th> Country of origin </Th> <Td> United States </Td> </Tr> <Tr> <Th> Original language (s) </Th> <Td> English </Td> </Tr> <Tr> <Th> No. of seasons </Th> <Td> </Td> </Tr> <Tr> <Th> No. of episodes </Th> <Td> 4 (list of episodes) </Td> </Tr> <Tr> <Th colspan="2"> Production </Th> </Tr> <Tr> <Th> Executive producer (s) </Th> <Td> <Ul> <Li> Matt Olmstead </Li> <Li> Jason T. Reed </Li> <Li> Jay Beattie </Li> <Li> Dan Dworkin </Li> <Li> Matthew Fernandes </Li> <Li> Arthur Spanos </Li> <Li> Vanessa Wong </Li> <Li> David Von Ancken </Li> </Ul> </Td> </Tr> <Tr> <Th> Running time </Th> <Td> 42 minutes </Td> </Tr> <Tr> <Th> Production company (s) </Th> <Td> <Ul> <Li> Dworkin / Beattie </Li> <Li> Brick Moon </Li> <Li> ABC Studios </Li> </Ul> </Td> </Tr> <Tr> <Th> Distributor </Th> <Td> Disney--ABC Domestic Television </Td> </Tr> <Tr> <Th colspan="2"> Release </Th> </Tr> <Tr> <Th> Original network </Th> <Td> ABC </Td> </Tr> <Tr> <Th> Original release </Th> <Td> April 2, 2018 (2018 - 04 - 02)--present (present) </Td> </Tr> <Tr> <Th colspan="2"> External links </Th> </Tr> <Tr> <Td colspan="2"> Website </Td> </Tr> </Table>''. Corresponding model predictions and associated probabilities are shown across different CMT methods. 'Context Token Prob.' is not reported since there is no context token to measure this on for irrelevant context.}
\label{tab:case-nq-irrelevant}
\end{table*}

%% file: Tables/case_by_case/druid.tex
\begin{table*}[ht]
\centering
\scriptsize
\begin{tabular}{lllc c r r r r}
\toprule
CMT &
Prediction &
\shortstack[l]{Prediction \\ w/ Context} &
BCU &
CCU &
\shortstack[l]{Prediction \\ Prob.} &
\shortstack[l]{Prediction Prob. \\ w/ Context} &
\shortstack[l]{Memory Token Prob. \\ w/ Context} &
\shortstack[l]{Context Token Prob. \\ w/ Context} \\
\midrule
Regular        & False & False & True  & 0.62  & 0.83 & 0.93 & 0.93 & 0.93 \\
Prompting     & False & False & True  & 0.62  & 0.83 & 0.93 & 0.93 & 0.93 \\
Fine-tuning   & False & False & True  & 0.61  & 0.87 & 0.92 & 0.92 & 0.92 \\
Multi-Agent   & False & False & True  & 0.62  & 0.83 & 0.93 & 0.93 & 0.93 \\
ACD           & False & False & True  & 0.46  & 0.83 & 0.91 & 0.91 & 0.91 \\
COIECD        & False & False & True  & 0.37  & 0.83 & 0.89 & 0.89 & 0.89 \\
PH3 +context  & False & False & True  & 0.83  & 0.83 & 0.97 & 0.97 & 0.97 \\
PH3 +memory   & False & None  & False & -0.74 & 0.83 & 0.66 & 0.21 & 0.21 \\
\bottomrule
\end{tabular}
\caption{\textbf{Claim:} ``HPV vaccine increases the risk of developing cervical cancer''. \textbf{Evidence (gold):} ``According to the World Health Organization, cervical cancer is the fourth most common cancer in women worldwide. In 2022, there were approximately 660,000 new cases of cervical cancer and around 350,000 deaths from cervical cancer. Infection by the sexually-transmitted human papillomavirus (HPV) is the main cause of cervical cancer. HPV vaccination can help prevent infection with high-risk HPV types, thereby reducing the risk of cervical cancer.''. Corresponding model predictions and associated probabilities are shown across different CMT methods.}
\label{tab:case-druid-gold}
\end{table*}
\begin{table*}[ht]
\centering
\scriptsize
\begin{tabular}{lllc c r r r r}
\toprule
CMT &
Prediction &
\shortstack[l]{Prediction \\ w/ Context} &
BCU &
CCU &
\shortstack[l]{Prediction \\ Prob.} &
\shortstack[l]{Prediction Prob. \\ w/ Context} &
\shortstack[l]{Memory Token Prob. \\ w/ Context} &
\shortstack[l]{Context Token Prob. \\ w/ Context} \\
\midrule
Regular        & False & False & False & 0.11 & 0.83 & 0.74 & 0.74 & 0.25 \\
Prompting      & False & False & False & 0.11 & 0.83 & 0.74 & 0.74 & 0.25 \\
Fine-tuning    & False & False & False & -0.33 & 0.87 & 0.90 & 0.90 & 0.00 \\
Multi-Agent    & False & None  & True  & 0.85 & 0.83 & 0.87 & 0.00 & 0.87 \\
ACD            & False & False & False & 0.04 & 0.83 & 0.79 & 0.79 & 0.20 \\
COIECD         & False & False & False & 0.05 & 0.83 & 0.79 & 0.79 & 0.20 \\
PH3 +context   & False & False & False & 0.03 & 0.83 & 0.80 & 0.80 & 0.19 \\
PH3 +memory    & False & None  & True  & 0.81 & 0.83 & 0.84 & 0.14 & 0.84 \\
\bottomrule
\end{tabular}
\caption{\textbf{Claim:} ``HPV vaccine increases the risk of developing cervical cancer''. \textbf{Evidence (conflicting):} ``Human papillomaviruses (HPV) are a group of sexually transmitted infections, some of which can increase the risk of cervical, anal, penile, vaginal, vulvar, and oropharyngeal cancers. Vaccines against HPV became available in 2006, and as the first adolescents who received the vaccine age, the incidence of these cancers is expected to decrease. Some experts believe that cervical cancer, which is associated with HPV in more than 90 percent of cases, could be virtually eliminated in the coming decades. [...] A recent study published in Cancer Epidemiology, Biomarkers \& Prevention investigated how the cervical cancer elimination timeline might differ between high-poverty and low-poverty communities. As the first adolescents who received the HPV vaccine are only beginning to enter the age range where cervical cancer is prominent, researchers have relied on statistical modeling to predict how vaccination will impact cervical cancer incidence in the coming years. In this study, Jennifer Spencer, PhD, an assistant professor at the University of Texas at Austin, who performed this work as a research fellow at the Harvard School of Public Health, and her colleagues assembled a series of models based on current data from counties in the highest and lowest quartile of poverty in the U.S.''. Corresponding model predictions and associated probabilities are shown across different CMT methods.}
\label{tab:case-druid-conflicting}
\end{table*}
\begin{table*}[ht!]
\centering
\scriptsize
\begin{tabular}{lllc c r r r r}
\toprule
CMT &
Prediction &
\shortstack[l]{Prediction \\ w/ Context} &
BCU &
CCU &
\shortstack[l]{Prediction \\ Prob.} &
\shortstack[l]{Prediction Prob. \\ w/ Context} &
\shortstack[l]{Memory Token Prob. \\ w/ Context} &
\shortstack[l]{Context Token Prob. \\ w/ Context} \\
\midrule
Regular        & False & None  & False & -0.88 & 0.87 & 0.90 & 0.09 & - \\
Prompting      & False & None  & False & -0.88 & 0.87 & 0.90 & 0.09 & - \\
Fine-tuning    & False & False & True  & -0.03 & 0.79 & 0.76 & 0.76 & - \\
Multi-Agent    & False & False & True  & -0.88 & 0.87 & 0.90 & 0.09 & - \\
ACD            & False & None  & False & -0.51 & 0.87 & 0.57 & 0.42 & - \\
COIECD         & False & None  & False & -0.46 & 0.87 & 0.53 & 0.46 & - \\
PH3 +context   & False & None  & False & -0.85 & 0.87 & 0.87 & 0.12 & - \\
PH3 +memory    & False & None  & False & -0.97 & 0.87 & 0.94 & 0.02 & - \\
\bottomrule
\end{tabular}
\caption{\textbf{Claim:} ``Claims ancient Greek philosopher Aristotle said, "The place where your talent meets the world’s needs is the job God has in mind for you.''. \textbf{Context (irrelevant):} ``Aristotle explains what he has in mind by comparing akrasia to the condition of other people who might be described as knowing in a way, but not in an unqualified way. His examples are people who are asleep, mad, or drunk; he also compares the akratic to a student who has just begun to learn a subject, or an actor on the stage (1147a10–24). All of these people, he says, can utter the very words used by those who have knowledge; but their talk does not prove that they really have knowledge, strictly speaking. [...] - Barney, Rachel, 2008, "Aristotle’s Argument for a Human Function", Oxford Studies in Ancient Philosophy, 34(Summer): 293–322. [...] - Russell, Daniel C., 2012a, "Aristotle’s Virtues of Greatness", Oxford Studies in Ancient Philosophy, Supplementary Volume 2012, Virtue and Happiness: Essays in Honour of Julia Annas, pp. 115–147.''. Corresponding model predictions and associated probabilities are shown across different CMT methods. 'Context Token Prob.' is not reported since there is no context token to measure this on for irrelevant context.}
\label{tab:case-druid-irrelevant}
\end{table*}

%% file: Tables/counterfact_memorisation.tex
\begin{table}[h]
    \centering
    \scriptsize
    \begin{tabular}{l l}
    \toprule
    Model     & Accuracy \\
    \midrule
        GPT-2 XL & 71.8 \\
        Pythia & 99.6 \\
        Qwen 1.5B & 77.0 \\
        Qwen 1.5B-I & 83.1 \\
        Qwen 7B & 79.7 \\
        Qwen 7B-I & 93.6 \\
        Qwen 32B & 78.0 \\
        Qwen 32B-I & 94.5 \\
        Command A & 90.6 \\
        GPT-4.1 mini & 89.4 \\
        GPT-4.1 & 97.7 \\
    \bottomrule
    \end{tabular}
    \caption{Accuracy, proxying memorisation rate, on samples from CounterFact without context.}
    \label{tab:counterfact-gold-memorisation}
\end{table}

%% file: Tables/tuned_prompts.tex
\begin{table}[h]
    \centering
    \scriptsize
    \setlength{\tabcolsep}{3.8pt}
    \begin{tabular}{l l l l}
    \toprule
        Dataset & Model & & Prompt \\
    \midrule
        CounterFact & {\scshape GPT2-XL} & 1.5B & \emph{default} \\
        & {\scshape Pythia} & 6.9B & Prompt \#10 (ChatGPT) \\
        & {\scshape Qwen2.5} & 1.5B & Prompt \#1 (\citet{jin-etal-2024-cutting}) \\
        & & 7B & Prompt \#11 (ChatGPT) \\
        & & 32B & Prompt \#8 (ChatGPT) \\
        & {\scshape Qwen2.5-I} & 1.5B & Instruct-prompt \#4 (manual) \\
        & & 7B & Instruct-prompt \#11 (ChatGPT) \\
        & & 32B & Instruct-prompt \#3 (manual) \\
        & {\scshape Command A} & 111B & Prompt \#5 (ChatGPT) \\
        & {\scshape GPT-4.1 mini} & & Prompt \#11 (ChatGPT) \\
        & {\scshape GPT-4.1} & & Prompt \#1 (\citet{jin-etal-2024-cutting}) \\
    \midrule
        NQ & {\scshape GPT2-XL} & 1.5B & Prompt \#2 (manual) \\
        & {\scshape Pythia} & 6.9B & \emph{default} \\
        & {\scshape Qwen2.5} & 1.5B & Prompt \#7 (ChatGPT) \\
        & & 7B & Prompt \#6 (ChatGPT) \\
        & & 32B & Prompt \#5 (manual) \\
        & {\scshape Qwen2.5-I} & 1.5B & Prompt \#5 (manual) \\
        & & 7B & Prompt \#3 (manual) \\
        & & 32B & \emph{default} \\
        & {\scshape Command A} & 111B & \emph{default} \\
        & {\scshape GPT-4.1 mini} & & Prompt \#3 (manual) \\
        & {\scshape GPT-4.1} & & Prompt \#3 (manual) \\
    \midrule
        DRUID & {\scshape GPT2-XL} & 1.5B & Prompt \#8 (ChatGPT) \\
        & {\scshape Pythia} & 6.9B & Prompt \#2 (manual) \\
        & {\scshape Qwen2.5} & 1.5B & Prompt \#2 (manual) \\
        & & 7B & Prompt \#11 (Microsoft Copilot) \\
        & & 32B & Prompt \#1 (manual) \\
        & {\scshape Qwen2.5-I} & 1.5B & \emph{default} \\
        & & 7B & \emph{default} \\
        & & 32B & Prompt \#2 (manual) \\
        & {\scshape Command A} & 111B & Prompt \#1 (manual) \\
        & {\scshape GPT-4.1 mini} & & Prompt \#6 (ChatGPT) \\
        & {\scshape GPT-4.1} & & Prompt \#5 (manual) \\
    \bottomrule
    \end{tabular}
    \caption{The tuned prompts for each LM. \emph{default} denotes that the original prompt template (seen in \Cref{app:data-collection}) worked best. ``-I'' denotes instruction-tuned model versions. The source of the prompt is indicated in parenthesis. For brevity, we refer the reader to our code repository (to be released upon publication) for the full prompts.}
    \label{tab:tuned-prompts}
\end{table}